\documentclass[twocolumn]{article}
\usepackage{geometry}
\usepackage{float}
\usepackage[utf8]{inputenc} 
\usepackage{amsmath}
\usepackage{amssymb}
\usepackage{amsthm}
\usepackage{mathtools}
\usepackage{graphicx}
\usepackage{booktabs}
\usepackage{enumitem}
\usepackage{listings} 
\usepackage{xcolor}
\usepackage{mdframed} 
\usepackage[pagebackref=true]{hyperref}
\usepackage{caption}
\usepackage{subcaption}
\usepackage{tikz}
\usepackage{pgfplots}
\usetikzlibrary{math}
\usetikzlibrary{calc}
\usetikzlibrary{decorations.markings}
\usetikzlibrary{backgrounds,trees,hobby}
\usepackage{standalone}
\usepackage[stable]{footmisc}

\definecolor{solColor}{rgb}{1,0.5,0}
\definecolor{otherColor}{gray}{0.7}
\definecolor{light-gray}{gray}{0.95} 
\lstnewenvironment{fileformat}
{
\lstset{
escapeinside={(*}{*)},
backgroundcolor=\color{light-gray}
}
}
{
\endquote
}
\BeforeBeginEnvironment{fileformat}{
\begin{mdframed}[backgroundcolor=light-gray, roundcorner=10pt,leftmargin=1, rightmargin=1, innerleftmargin=15, innertopmargin=15,innerbottommargin=15, outerlinewidth=1, linecolor=light-gray]
}
\AfterEndEnvironment{fileformat}{
\end{mdframed} 
}

\newtheorem{theorem}{Theorem}

\newtheorem{definition}[theorem]{Definition}

\DeclareMathOperator{\argmax}{argmax}
\newcommand{\R}{\mathbb{R}}

\newcommand{\Z}{\mathbb{Z}}
\newcommand{\N}{\mathbb{N}}
\newcommand{\la}{\langle}
\newcommand{\ra}{\rangle}
\DeclarePairedDelimiter\abs{\lvert}{\rvert}

\mathchardef\mhyphen="2D

\title{Structured Prediction Problem Archive}

\author{Paul Swoboda$^{\circledcirc,\dagger}$, Bjoern Andres$^{\triangleleft}$, Andrea Hornakova$^\dagger$, Florian Bernard$^{\star}$, Jannik Irmai$^\triangleleft$, \\ Paul Roetzer$^{\ddagger,\star}$, Bogdan Savchynskyy$^{\diamond}$, David Stein$^\triangleleft$, Ahmed Abbas$^\dagger$}
\date{%
$^\circledcirc$Heinrich-Heine~University~D\"usseldorf, %
$^\dagger$MPI~for~Informatics, %
$^\ddagger$TU~Munich, %
$^\star$University~of~Bonn, %
$^\diamond$Heidelberg~University %
$^\triangleleft$TU~Dresden %
}

\begin{document}
\maketitle

\begin{abstract}
Structured prediction problems are one of the fundamental tools in machine learning.
In order to facilitate algorithm development for their numerical solution, we collect in one place a large number of datasets in easy to read formats for a diverse set of problem classes.
We provide archival links to datasets, description of the considered problems and problem formats, and a short summary of problem characteristics including size, number of instances etc.
For reference we also give a non-exhaustive selection of algorithms proposed in the literature for their solution.
We hope that this central repository will make benchmarking and comparison to established works easier.
    We welcome submission of interesting new datasets and algorithms for inclusion in our archive\footnote{Corresponding email address \href{mailto:paul.swoboda@hhu.de}{paul.swoboda@hhu.de}}.
\end{abstract}

\tableofcontents

\clearpage


\section{Structured Prediction}
Structured prediction is the task to predict structured solutions, which, in the following, will mean vectors that are subject to a number of constraints.
In particular, we restrict ourselves to integer linear programs, where all variables are binary subject to a number of constraints which commonly can be written as linear (in-)equalities, see Section~\ref{sec:ilp} below.
Due to the constraint structure finding even a feasible solution can be NP-hard. 
But even if the constraint structure admits easy construction of solutions, finding the best one w.r.t.\ an objective (often called energy in the literature) is typically NP-hard.
This and the large problem sizes typically occurring in machine learning and computer vision make structured prediction an interesting and difficult algorithmic challenge.

Over the years a large body of research has been devoted to efficient and scalable approximative and also optimal solvers for special subclasses of problems.
Somewhat less attention has been paid to generic, that is problem agnostic solvers.
In order to assess the empirical performance, datasets of instances in various problem classes have been  used.
It is the goal of the Structured Prediction Problem Archive to present and preserve interesting datasets coming from a large number of different problem classes together with references to algorithms which were proposed for their solution.
Unfortunately, in many previous works experiments were done on differing subsets of datasets, making comparison across different algorithms harder than it should be.
Similar projects have been pursued for Markov Random Fields~\cite{kappes2015comparative,szeliski2008comparative} with additional algorithm evaluations.

We hope that our project will make empirical testing of algorithms easier.
Our work provides a single point of reference from which problem instances can be obtained.
We hope that this will encourage algorithm testing on large corpora of instances.

\subsection{Integer Linear Programs}
\label{sec:ilp}
All structured prediction problems we collect can be written as integer linear programs (ILP).
An ILP is a linear minimization problem over integral variables that are subject to a number of linear constraints.
We constrain ourselves to $\{0,1\}$-valued variables.
Given an objective vector $c \in \R^n$, a constraint matrix $A \in \R^{n \times m}$ and a constraint right hand side $b \in \R^m$ we can write an ILP as

\begin{equation}
    \tag{ILP}
    \begin{array}{rl}
    \min_{x \in \{0,1\}^n}
    & \sum_{i\in[n]} c_i x_i \\
    \text{s.t.}
    & Ax \leq b \\
    \end{array}
\end{equation}
For greater flexibility we allow also equality constraints, i.e.\  $Ax = b$ or a mixture of inequalities and equalities.

\subsection{ILP File Format}
\label{sec:ilp-file-format}
Whenever applicable, e.g.\ when the number of constraints can be polynomially bounded, we store the problems in an LP file format that standard ILP solvers can read.
The format is structured as follows:

{\small
\begin{fileformat}
Minimize
(*$\sum\limits_i c_i var_i$*)
Subject to
ineq_id(*${}_1$*): (*$\sum\limits_{i}$*) (*$a_{1i} var_{i}$*)  (*$\{\leq|\geq|=\}$*) (*$b_1$*)
.
.
.
ineq_id(*${}_m$*): (*$\sum\limits_{i}$*) (*$a_{mi} var_{i}$*)  (*$\{\leq|\geq|=\}$*) (*$b_m$*)
Bounds
Binaries
(*$var_1$*)
.
.
.
(*$var_n$*)
End
\end{fileformat}
}

\subsection{Algorithms}
\begin{description}[style=unboxed]
\item[CPLEX~\cite{cplex}:] Classic leading commercial ILP solver from IBM.
\item[Gurobi~\cite{gurobi}:] Newer leading commercial ILP solver.
\item[Mosek~\cite{mosek}:] Another leading commercial ILP solver.
\item[SCIP~\cite{achterberg2009scip}:] Leading academic ILP and constraint integer solver.
\item[BDD Min-Marginal Averaging~\cite{lange2021efficient}:]
Decompose ILPs into subproblems represented by binary decision diagrams. Solve the resulting Lagrange decomposition by a sequential min-marginal averaging or with a parallel extension.
\item[Fast Discrete Optimization on GPU (1astDOG)~\cite{abbas2022fastdog}:] Extension of the BDD min-marginal averaging to a massively parallel GPU solver.
\end{description}

\clearpage
\section{Markov Random Fields}
\label{sec:mrf}
Markov Random Fields (MRF) are a basic model for obtaining structured probability distributions that factorize according to a graph.
Finding the most probable configuration is called maximum-a-posteriori inference or MAP-MRF for short.
Here we consider only discrete MRFs, i.e.\ random variables of the MRF can only take a finite number of values.
MAP-MRF for discrete MRFs can be cast as an ILP~\cite{werner2007linear}.

\begin{definition}[MRF]
    \label{def:mrf}
Let an undirected graph $G=(V,E)$, label spaces $\mathcal{L}_v = \{1,\ldots, \abs{\mathcal{L}_v}\}$ for each variable $v \in V$ and $\mathcal{L} = \prod_{v \in V} \mathcal{L}_v$ and 
unary potentials $\phi_v : \mathcal{L}_v \rightarrow \R$, $v \in V$ as well as pairwise potentials
$\phi_{uv} : \mathcal{L}_u \times \mathcal{L}_v \rightarrow \R$, $uv \in E$ be given.
The probability distribution defined by an MRF $(V,E,\mathcal{L},\phi)$ is given by
\begin{equation}
\mathbb{P}_{\phi}(x) = \frac{\prod_{v \in V} \phi_v(x_v) \cdot \prod_{uv \in E} \phi_{uv}(x_u,x_v)}{ \underbrace{\sum_{x \in \mathcal{L}} \prod_{v \in V} \phi_v(x_v) \cdot \prod_{uv \in E} \phi_{uv}(x_u,x_v)}_{ := Z } } \,.
\end{equation}
The normalization constant $Z$ is called the partition function.
\end{definition}

\begin{definition}[MAP]
The maximum-a-posteriori element is defined as
\begin{equation}
\label{eq:map-mrf}
x^* \in \argmax_{x \in \mathcal{L}} \mathbb{P}(x)\,.
\end{equation}
After taking negative logarithms $\theta_v = - \log(\phi_v)$ $\forall v \in V$ and $\theta_{uv} = -\log(\phi_{uv})$ $\forall uv \in E$ the MAP problem~\eqref{eq:map-mrf} can be written as
\begin{equation}
\min_{x \in \mathcal{L}} \theta(x) = \sum_{v \in V} \theta_v(x_v) + \sum_{uv \in E} \theta_{uv}(x_u,x_v) \,.
\end{equation}
\end{definition}

\begin{figure}[H]
    \centering
    \begin{tikzpicture}
\foreach \x in {0,...,2} {
	\foreach \y in {0,...,2} {
		\foreach \l in {0,...,2} {
			\tikzmath{\h = (\l-1)/5;}
			\node[circle,draw=black,ultra thin,fill=otherColor,scale=0.3] (\x-\y-\l) at (\x + \h + 0.5,\y + \h + 0.5) {}; 
		}
			\draw[black,rotate around={45:(\x + 0.5,\y + 0.5)}] (\x+ 0.5-0.40,\y+ 0.5-0.07) rectangle (\x+ 0.5+0.40,\y+0.5+0.07);
	}
}
\foreach \x in {0,...,2} {
	\tikzmath{\xn = \x+1;}
	\foreach \y in {0,...,2} {
		\tikzmath{\yn = \y+1;}
		\foreach \l in {0,...,2} {
			\tikzmath{\hpp = (\l-3)/5;}
			\tikzmath{\hp = (\l-2)/5;}
			\tikzmath{\h = (\l-1)/5;}
			\tikzmath{\hn = (\l)/5;}
			\tikzmath{\hnn = (\l+1)/5;}
			\ifnum \x < 2
				\draw [ultra thin, otherColor] (\x + \h + 0.5,\y + \h + 0.5) -- (\xn + \h + 0.5,\y + \h + 0.5);
				\ifnum \l=0
					\draw [ultra thin, otherColor] (\x + \h + 0.5,\y + \h + 0.5) -- (\xn + \hn + 0.5,\y + \hn + 0.5);
					\draw [ultra thin, otherColor] (\x + \h + 0.5,\y + \h + 0.5) -- (\xn + \hnn + 0.5,\y + \hnn + 0.5);
				\fi
				\ifnum \l=1
					\draw [ultra thin, otherColor] (\x + \h + 0.5,\y + \h + 0.5) -- (\xn + \hp + 0.5,\y + \hp + 0.5);
					\draw [ultra thin, otherColor] (\x + \h + 0.5,\y + \h + 0.5) -- (\xn + \hn + 0.5,\y + \hn + 0.5);
				\fi
				\ifnum \l=2
					\draw [ultra thin, otherColor] (\x + \h + 0.5,\y + \h + 0.5) -- (\xn + \hp + 0.5,\y + \hp + 0.5);
					\draw [ultra thin, otherColor] (\x + \h + 0.5,\y + \h + 0.5) -- (\xn + \hpp + 0.5,\y + \hpp + 0.5);
				\fi 
			\fi
			\ifnum \y < 2
				\draw [ultra thin, otherColor] (\x + \h + 0.5,\y + \h + 0.5) -- (\x + \h + 0.5,\yn + \h + 0.5);
				\ifnum \l=0
				\draw [ultra thin, otherColor] (\x + \h + 0.5,\y + \h + 0.5) -- (\x + \hn + 0.5,\yn + \hn + 0.5);
				\draw [ultra thin, otherColor] (\x + \h + 0.5,\y + \h + 0.5) -- (\x + \hnn + 0.5,\yn + \hnn + 0.5);
				\fi
				\ifnum \l=1
				\draw [ultra thin, otherColor] (\x + \h + 0.5,\y + \h + 0.5) -- (\x + \hp + 0.5,\yn + \hp + 0.5);
				\draw [ultra thin, otherColor] (\x + \h + 0.5,\y + \h + 0.5) -- (\x + \hn + 0.5,\yn + \hn + 0.5);
				\fi
				\ifnum \l=2
				\draw [ultra thin, otherColor] (\x + \h + 0.5,\y + \h + 0.5) -- (\x + \hp + 0.5,\yn + \hp + 0.5);
				\draw [ultra thin, otherColor] (\x + \h + 0.5,\y + \h + 0.5) -- (\x + \hpp + 0.5,\yn + \hpp + 0.5);
				\fi 
			\fi
		}
	}
}
\tikzmath{\hd = (1)/5;}
\node[circle,draw=black,ultra thin,fill=orange,scale=0.3] (x1) at (0.5,0.5) {}; 
\node[circle,draw=black,ultra thin,fill=orange,scale=0.3] (x2) at (1.5-\hd,0.5-\hd) {}; 
\node[circle,draw=black,ultra thin,fill=orange,scale=0.3] (x3) at (2.5+\hd,0.5+\hd) {}; 
\node[circle,draw=black,ultra thin,fill=orange,scale=0.3] (x4) at (0.5,1.5) {}; 
\node[circle,draw=black,ultra thin,fill=orange,scale=0.3] (x5) at (1.5+\hd,1.5+\hd) {}; 
\node[circle,draw=black,ultra thin,fill=orange,scale=0.3] (x6) at (2.5,1.5) {}; 
\node[circle,draw=black,ultra thin,fill=orange,scale=0.3] (x7) at (0.5+\hd,2.5+\hd) {}; 
\node[circle,draw=black,ultra thin,fill=orange,scale=0.3] (x8) at (1.5,2.5) {}; 
\node[circle,draw=black,ultra thin,fill=orange,scale=0.3] (x9) at (2.5+\hd,2.5+\hd) {}; 
\path [draw, orange] (x1) -- (x2) node[midway] (12) {};
\path [draw, orange] (x2) -- (x3) node[midway] (23) {};
\path [draw, orange] (x1) -- (x4) node[midway] (14) {};
\path [draw, orange] (x2) -- (x5) node[midway] (25) {};
\path [draw, orange] (x3) -- (x6) node[midway] (36) {};
\path [draw, orange] (x4) -- (x5) node[midway] (45) {};
\path [draw, orange] (x5) -- (x6) node[midway] (56) {};
\path [draw, orange] (x4) -- (x7) node[midway] (47) {};
\path [draw, orange] (x5) -- (x8) node[midway] (58) {};
\path [draw, orange] (x6) -- (x9) node[midway] (69) {};
\path [draw, orange] (x7) -- (x8) node[midway] (78) {};
\path [draw, orange] (x8) -- (x9) node[midway] (89) {};
\node[orange, left of=47, node distance = 6pt] (){};
\node[orange, left of=58, node distance = 6pt] (){};
\node[orange, left of=36, node distance = 6pt] (){};
\node[orange, below of=12, node distance = 6pt] (){};
\path [draw, solColor] (x1) -- (x2) node[midway] (12) {};
\node[solColor, below of=12, node distance = 12pt] (){\tiny \textcolor{black}{$\theta_{ab}(2, 1)$}};
\path [draw, solColor] (x2) -- (x3) node[midway] (23) {};
\node[solColor, below of=23, node distance = 14pt] (){\tiny \textcolor{black}{$\theta_{bc}(1, 3)$}};
\path [draw, solColor] (x7) -- (x8) node[midway] (78) {};
\node[solColor, above of=78, node distance = 10pt] (){\tiny \textcolor{black}{$\theta_{gh}(3, 2)$}};
\end{tikzpicture}
    \caption{
    An exemplary 3-label MRF on a grid graph.
    A selected solution is indicated by the orange nodes and edges.
    }
    \label{fig:mrf-illustration}
\end{figure}
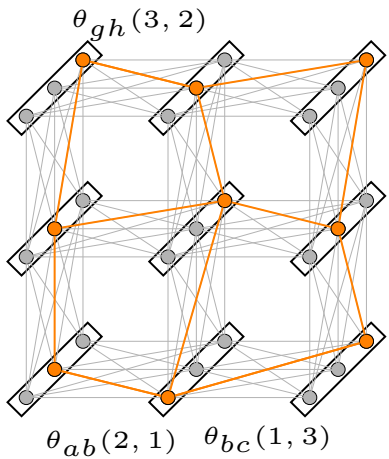

For an illustration of an exemplary MRF see Figure~\ref{fig:mrf-illustration}.

\subsection{File Format}
\label{sec:mrf-file-format}
We use an extension of the UAI file format~\url{https://www.cs.huji.ac.il/project/PASCAL/fileFormat.php}.

{
\small
\begin{fileformat}
MARKOV
(*$\abs{V}$*)
(*$\abs{\mathcal{L}_1}$*) ... (*$\abs{\mathcal{L}_{\abs{V}}}$*) 
(*$\abs{V} + \abs{E}$*)
1 1
.
.
.
1 (*$\abs{V}$*) 
2 (*$i_1$*) (*$j_1$*) 
.
.
.
2 (*$i_{\abs{E}}$*) (*$j_{\abs{E}}$*) 

(*$\abs{\mathcal{L}_1}$*) 
(*$\theta_1(1)$*) ... (*$\theta_1(\abs{\mathcal{L}_1})$*)
.
.
.
(*$\abs{\mathcal{L}_{\abs{V}}}$*)
(*$\theta_{\abs{V}}(1)$*) ... (*$\theta_{\abs{V}}(\abs{\mathcal{L}_{\abs{V}}})$*)

(*$\abs{\mathcal{L}_{i_1}} \cdot \abs{\mathcal{L}_{j_1}} $*) 
(*$\theta_{i_1 j_1}(1,1)$*) (*$\theta_{i_1 j_1}(1,2)$*) ... (*$\theta_{i_1 j_1}(\abs{\mathcal{L}_{i_1}}, \abs{\mathcal{L}_{j_1}})$*)
.
.
.
(*$\abs{\mathcal{L}_{i_{\abs{E}}}} \cdot \abs{\mathcal{L}_{j_{\abs{E}}}} $*) 
(*$\theta_{i_{\abs{E}} j_{\abs{E}}}(1,1)$*) ... (*$\theta_{i_{\abs{E}} j_{\abs{E}}}(\abs{\mathcal{L}_{i_{\abs{E}}}}, \abs{\mathcal{L}_{j_{\abs{E}}}})$*)
\end{fileformat}
}
The file is structured as follows:
First comes the preamble (MARKOV, line 1),
then the number of variables ($\abs{V}$, line 2),
then the number of labels for each variable in order,
then the number of unary and pairwise potentials.
Next come the index lines:
for each unary and pairwise potential the number of nodes (1 for variables, 2 for edges) followed by the variable indices.
Last come function tables in the order given by the index lines.
Each function table gives the number of entries in the function table $\abs{\mathcal{L}_i}$ for variables $i$ and $\abs{\mathcal{L}_i} \cdot \abs{\mathcal{L}_j}$ for edge $ij$ followed by the values of the corresponding potentials $\theta_i$ or $\theta_{ij}$ respectively.

Additionally we provide LP-files for all datasets using the local polytope relaxation~\cite{werner2007linear} for general pairwise potentials and a compactified one equivalent to the local polytope for Potts potentials.

\subsection{Datasets}
The below datasets object-seg, color-seg, color-seg-n4, color-seg-n8 and protein-folding are part of the OpenGM benchmark~\cite{kappes2015comparative}.
These datasets are additionally given in the OpenGM hdf5 format.

\subsubsection[object-seg]{object-seg\footnote{\url{https://keeper.mpdl.mpg.de/f/716a0a8621f44d0ba9be/?dl=1}}}
Segmentation of objects in 3 images of size up to 424720 nodes with up to 4 semantic classes and Potts pairwise potentials on an 8-neighborhood.

\subsubsection[color-seg]{color-seg\footnote{\url{https://keeper.mpdl.mpg.de/f/cdcae8287a36403e9ecd/?dl=1}}}
Image segmentation with up to 424720 nodes on an 8-neigborhood with up to 4 labels. 
Pairwise potentials are of Potts type.
Instances are originally from~\cite{alahari2009dynamic}.

\subsubsection[color-seg-n4]{color-seg-n4\footnote{\url{https://keeper.mpdl.mpg.de/f/76b5ea50eba24c1d9111/?dl=1}}}
Image segmentation of up to 76800 nodes on a 4-neighborhood with up to 12 labels. 
Pairwise potentials are of Potts type.
Instances are originally from~\cite{lellmann2011continuous}.

\subsubsection[color-seg-n8]{color-seg-n8\footnote{\url{https://keeper.mpdl.mpg.de/f/0077922f8e054fee8860/?dl=1}}}
Image segmentation of up to 76800 nodes on a 8-neighborhood with up to 12 labels. 
Pairwise potentials are of Potts type.
Instances are originally from~\cite{lellmann2011continuous}.

\subsubsection[protein-folding]{protein-folding\footnote{\url{https://keeper.mpdl.mpg.de/f/927e7cfa31fa4cd9bcd4/?dl=1}}}
29 instances of moderate size with sparse and full connectivity and pairwise potentials of general type.
Instances are originally from~\cite{jaimovich2006towards,elidan2011probabilistic} for solving the protein folding problem.

\subsubsection[Global 6D Object Pose Estimation]{Global 6D Object Pose Estimation\footnote{\url{https://keeper.mpdl.mpg.de/f/399de399aa1146c29556/?dl=1}}}
$32$ problem instances with $600-4800$ variables each and $13$ labels per node. The models are all fully connected.
Instances are cropped versions of those from~\cite{michel2017global} for the 6D object pose estimation problem.
The crops are given by masks covering the object of interest and reduce problem size.
We do not provide ILP-files in the format of Section~\ref{sec:ilp-file-format} due to the large size of the resulting encoding.

\subsection{Algorithms}
\begin{description}[style=unboxed]
\item[OpenGM~\cite{kappes2015comparative}:] Library of $\sim$ 50 different algorithms for inference in MRFs.
\item[Optimization Monograph for MRFs~\cite{savchynskyy2019discrete}] Monograph on optimization techniques (Lagrange decomposition, message passing, subgradient methods, max-flow) for MRFs.
\item[Tree-reweigthed sequential message passing (TRW-S)~\cite{kolmogorov2005convergent}:] Decompose the graph into chains and sequentially visit nodes and perform min-marginal averaging reusing previous computations. A lower bound is monotonically increasing. Rounding is performed by sequentially fixing visited variables to their locally best label.
\item[Sequential Reweighted Message Passing Revisited (SRMP)~\cite{kolmogorov2014new}:] Extension of TRW-S~\cite{kolmogorov2005convergent} to higher order MRFs.
\item[Boykov-Kolmogorov max-flow~\cite{boykov2004experimental}:] Convert $2$-label MRFs with submodular potentials to max-flow problems and solve them with an efficient combinatorial max-flow solver.
\item[Quadratic Binary Program Optimization (QPBO)~\cite{kolmogorov2007minimizing}:] Convert arbitrary $2$-label MRFs to an extended maximum flow problem and solve the resulting relaxation with~\cite{boykov2004experimental}.
\item[Quadratic Binary Program Optimization Improved (QPBOi)~\cite{rother2007optimizing}:] Extension of QPBO~\cite{kolmogorov2007minimizing} to allow for probing variables and fixing them even if QPBO cannot determine its value.
\item[$\alpha$-expansion~\cite{boykov2001fast}:] Given a multi-label Potts MRF fix a label and compute best move that can pick the fixed label via reduction to maximum-flow. Iterate over all labels until convergence.
\item[Local Submodular Approximation (LSA)~\cite{gorelick2014submodularization}:] Local submodular approximation of the binary MRF energies solved iteratively with max-flow.
\item[MQPBO~\cite{kohli2008partial}:] Transformation of ordered multi-label MRFs to an extended binary MRF.
\item[Max-Product Message Passing (MPLP)~\cite{globerson2007fixing}:] Message passing on a decomposition into edge subproblems.
\item[Subproblem Tree Calibration~\cite{wang2013subproblem}:] A unified view of message passing on different subproblems.
\item[Reduction to Perfect Matching~\cite{schraudolph2010polynomial}:] Solve planar binary pairwise MRFs with unary potentials only on the outer boundary exactly by transforming to minimum cost perfect matching. For arbitrary planar binary MRFs transform to minimum cost perfect matching with additional Lagrange multipliers.
\item[Planar Cycle Covering Graphs~\cite{yarkony2011planar}:] Improved transformation of planar binary MRFs to perfect matching with additional Lagrange multipliers.
\item[Tree Block Coordinate Ascent~\cite{sontag2009tree}:] Decompose MRF into tree subproblems and compute Lagrange multiplier updates for full trees simultaneously.
\item[Minorant Averaging~\cite{shekhovtsov2016solving}:] Iteratively compute minorants of chain subproblems in parallel on GPU and average them.
\item[MPLP++~\cite{tourani2018mplp++}:] Modification of the MPLP~\cite{globerson2007fixing} with more effective update steps leading to fast convergence for dense subproblems.
\item[SPAM~\cite{tourani2020taxonomy}:] Adaptive subproblem construction and large update steps leading to fast convergence for arbitrary graph structures.
\item[Adaptive Diminishing Smoothing~\cite{savchynskyy2012efficient}:] Solve smoothed dual of local polytope relaxation with an adaptively chosen smoothing parameter.
\item[Frustrated Cycle Search~\cite{sontag2012efficiently}:] Tighten the relaxation used in MPLP~\cite{globerson2007fixing} with triplet potentials coming from violated cycle inequalities.
\item[Augmenting DAG~\cite{werner2007linear}:] Find sequences of message passing operations that will increase lower bound along so-called augmenting directed acyclic subgraphs.
\item[CombiLP~\cite{savchynskyy2013global}:] Splitting MRF into easy/hard part and solving the easy one with efficient TRWS~\cite{kolmogorov2005convergent} and the hard one with an ILP solver and recombining the respective solutions into a globally optimal one.
\item[Dense CombiLP~\cite{haller2018exact}:] Extension of CombiLP with better splitting criterion into easy/hard part resulting in an easier hard part. 
\item[Subgradient on Dual Decomposition~\cite{komodakis2007mrf}:] Lagrange decomposition into trees optimized with subgradient ascent.
\item[Bundle Method~\cite{kappes2012bundle}:] Bundle method optimizing a Lagrange decomposition into trees.
\item[Frank-Wolfe Bundle Method~\cite{swoboda2019map}:] Lagrange decomposition into minimal number of tree subproblems solved with a bundle method using the Frank-Wolfe algorithm.
\item[Iterated Conditional Modes (ICM)~\cite{besag1986statistical}:] Iteratively improve the labeling of a single variable and keep the others fixed.
\item[Fixed-point iteration~\cite{leordeanu2009integer}:] Climbing and convergence guaranteeing discretization scheme.
\item[Lazy Flipper~\cite{andres2012lazy}:] ICM~\cite{besag1986statistical} extension for flipping assignments of $k$ variables at a time efficiently.
\item[Nesterov's scheme~\cite{savchynskyy2011study}:] Nesterov's fast gradient based algorithm on a smoothed approximate of the Lagrange dual.
\item[Primal-Dual~\cite{schmidt2011evaluation}:] A first order primal-dual algorithm optimizing the local polytope relaxation.
\item[Dead End Elimination (DEE) Persistency~\cite{desmet1992dead}:] Compute for every variable assignment whether changing the assignment to some other label always improves the energy. If so, elimiate variable/label assignment.
\item[Kovtun's persistency criterion~\cite{desmet1992dead}:] Solve auxiliary max-flow subproblems to compute certificates whether variable assignments are optimal.
\item[Improving Mapping Persistency~\cite{shekhovtsov2014maximum,shekhovtsov2015maximum,shekhovtsov2016higher}:] Find optimal variable assignments by constructing improving mapping that fixes optimal variabl assignments.
\item[Iterative Pruning Persistency~\cite{swoboda2013partial,swoboda2014partial}:] Iteratively shrink the MRF by removing nodes where persistency cannot be proved and modify the pruned problem with boundary terms coming from the deleted variables.
\end{description}

\clearpage
\section{Multicut}
\label{sec:multicut}

The multicut problem~\cite{chopra1993partition} (also known as correlation clustering~\cite{demaine2006correlation}) is to cluster graph nodes based on edge preferences.

\begin{definition}[Multicut]

Given an undirected weighted graph $G=(V,E,c)$ the multicut problem is to find a partition $\Pi = (\Pi_1,\ldots,\Pi_k)$ such that $\Pi_i \cap \Pi_j = \varnothing$ for $i \neq j$ and $\dot{\cup}_{i\in[k]} \Pi_j = V$ where $k$ is computed as part of the optimization process.
The \emph{cut} $\delta(\Pi_1,\ldots,\Pi_k)$ induced by a decomposition is the subset of those edges that straddle distinct clusters.

The space of all multicuts is denoted as
\begin{equation}
  \mathcal{M}_G = \left\{ \delta(V_1,\ldots,V_k) : 
\begin{array}{c}
  k \in \N \\
  V_1 \dot\cup \ldots \dot\cup V_k = V
\end{array} \right\}\,.
\end{equation} 

The minimum cost multicut problem is
\begin{equation}
  \label{eq:multicut}
  \min_{y \in \mathcal{M}_G} \la c, x \ra\,.
\end{equation} 
\end{definition}

\begin{figure}[H]
    \centering
    \scalebox{0.7}{ \definecolor{mycolor1}{HTML}{009900}
\tikzstyle{cut-edge}=[red,dashed,thick]
\tikzstyle{vertex}=[circle, draw, fill=white, inner sep=0pt, minimum width=1ex]
\tikzset{every picture/.append style={baseline,scale=1.1}}
\begin{tikzpicture}[scale=1.5]
\draw[draw=mycolor1!30, fill=mycolor1!30] plot[smooth cycle, tension=0.5] coordinates
{(-0.3, 2.3) (2.3, 2.3) (2.3, 0.7) (0.7, 0.7) (0.7, 1.7) (-0.3, 1.7)};
\draw[draw=mycolor1!30, fill=mycolor1!30] plot[smooth cycle, tension=0.5] coordinates
{(-0.3, -0.3) (-0.3, 1.3) (0.3, 1.3) (0.3, 0.3) (1.3, 0.3) (1.3, -0.3)};
\draw[draw=mycolor1!30, fill=mycolor1!30] plot[smooth cycle, tension=0.5] coordinates
{(1.7, -0.3) (1.7, 0.3) (2.7, 0.3) (2.7, 2.3) (3.3, 2.3) (3.3, -0.3)};
\draw (0, 0) -- (0, 1);
\draw (0, 0) -- (1, 0);
\draw (0, 2) -- (1, 2);
\draw (1, 1) -- (1, 2);
\draw (1, 1) -- (2, 1);
\draw (1, 2) -- (2, 2);
\draw (2, 1) -- (2, 2);
\draw (2, 0) -- (3, 0);
\draw (3, 0) -- (3, 1);
\draw (3, 1) -- (3, 2);
\draw[style=cut-edge] (0, 1) -- (0, 2);
\draw[style=cut-edge] (0, 1) -- (1, 1);
\draw[style=cut-edge] (1, 0) -- (1, 1);
\draw[style=cut-edge] (1, 0) -- (2, 0);
\draw[style=cut-edge] (2, 0) -- (2, 1);
\draw[style=cut-edge] (2, 1) -- (3, 1);
\draw[style=cut-edge] (2, 2) -- (3, 2);
\node[style=vertex] at (0, 0) {};
\node[style=vertex] at (1, 0) {};
\node[style=vertex] at (0, 1) {};
\node[style=vertex] at (0, 2) {};
\node[style=vertex] at (1, 1) {};
\node[style=vertex] at (1, 2) {};
\node[style=vertex] at (2, 1) {};
\node[style=vertex] at (2, 2) {};
\node[style=vertex] at (2, 0) {};
\node[style=vertex] at (3, 0) {};
\node[style=vertex] at (3, 1) {};
\node[style=vertex] at (3, 2) {};
\end{tikzpicture} }
    \caption{
    An exemplary multicut on a grid graph.
    The multicut consists of three partitions induced by the cut edges (red, dashed).
    }
    \label{fig:multicut-illustration}
\end{figure}
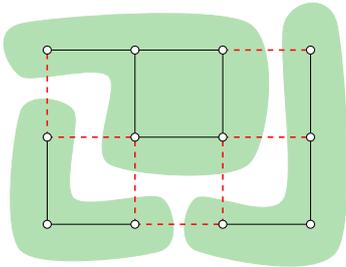

A multicut illustration can be seen in Figure~\ref{fig:multicut-illustration}

\subsection{File Format}
The multicut problem is given in the text file format as follows:

\begin{fileformat}
MULTICUT
(*$i_1$*) (*$j_1$*) (*$c_1$*)
.
.
.
(*$i_m$*) (*$j_m$*) (*$c_m$*)
\end{fileformat}
where $E = \{i_1 j_1, \ldots, i_m j_m\}$ and $c_1,\ldots,c_m$ are the corresponding edge weights.

\subsection{Benchmark datasets}

\subsubsection{CREMI}
The CREMI-challenge~\cite{cremi} is to group voxels from 3D-volumes of fruit-fly brain matter together whenever they belong to the same neuron.
The raw image data was acquired by~\cite{zheng2018complete} and converted to multiple multicut instances as detailed below.

\paragraph{Superpixel\footnote{\url{https://keeper.mpdl.mpg.de/f/811b88d4c97644d39ea9/?dl=1}}}
For converting the data into multicut instances the authors of~\cite{pape2017solving} first created super-pixels and then computed affinities between these for estimating probabilities that superpixels belong to the same neuron.
Instances are different crops of one global problem.
There are 3 small ($400000-600000$ edges), 3 medium ($4-5$ million edges) and 5 large ($28-650$ million edges) multicut instances.

\paragraph{Raw\footnote{\url{https://keeper.mpdl.mpg.de/f/3916d2da6aa840139206/?dl=1}}}
Multicut instances are derived directly from the voxel grid without conversion to superpixels.
Three test volumes sample A+, B+ and C+ from~\cite{cremi} were used.
Edge weights are computed by~\cite{torch_EM}.
There are two types of instances:
(i)~The three full problems where the underlying volumes have size $1250 \times 1250 \times 125$ with around $700$ million edges and
(ii)~six cropped problems created by halving each volume and creating the corresponding multicut instances each containing almost $340$ million edges.

\subsubsection[Cityscapes Instance Segmentation]{Cityscapes Instance Segmentation\footnote{\url{https://keeper.mpdl.mpg.de/f/80686a004ff84d96aaeb/?dl=1}}}

Unsupervised image segmentation on $59$ high resolution images ($2048 \times 1024$) taken from the validation set~\cite{cordts2016cityscapes}.
Conversion to multicut instances is done by computing the edge affinities produced by~\cite{abbas2021combinatorial} on a grid graph with $4$-connectivity and additional coarsely sampled longer range edges.
Each instance contains approximately $2$ million nodes and $9$ million edges. 

\subsubsection{OpenGM}
The OpenGM benchmark~\cite{kappes2015comparative} contains instances of the multicut problem from several applications.
These instances are stored in a more general graphical model format. 
For convenience we provide these instances here in the multicut file format as described above.

\paragraph{Natural image segmentation\footnote{\url{https://keeper.mpdl.mpg.de/f/7af922a52a354896a5d7/?dl=1}}}
Instances of the multicut problem are constructed from natural images from the Berkley segmentation dataset~\cite{MartinFTM01} as described in~\cite{andres2011closedness}. 
First, superpixels are computed by a watershed over-segmentation and a graph is constructed with one node for each superpixel and edges between adjacent superpixels.
Then, edge costs are estimated by a random forest that is trained on manually annotated data.
There are 100 instances with $156-3764$ nodes and  $439-10970$ edges.

\paragraph*{Knott3D\footnote{\url{https://keeper.mpdl.mpg.de/f/822134874a8f4c2aa0a9/?dl=1}}}
Instance of the multicut problem are constructed from volume images of an adult mouse somatosensory cortex, acquired by~\cite{knott2008serial}, as described in~\cite{andres2012globally}.
For this, superpixel graphs are computed with a combination of a random forest trained on manually annotated data and a watershed over-segmentation.
Edge costs are also estimated by a random forest.
There are three types of instances corresponding to different volume sizes.
In total there are 24 instances with $571-17073$ nodes and $3381-107060$ edges.

\paragraph*{Modularity clustering\footnote{\url{https://keeper.mpdl.mpg.de/f/33bb96c32ae04395a1e7/?dl=1}}}
Modularity is a measure for quality of a clustering of the nodes of an undirected graph.
Modularity clustering is the problem of finding a clustering of maximal modularity.
The modularity clustering problem can be formulated as a multicut problem on a complete graph~\cite{brandes2008modularity}.
The OpenGM benchmark contains instances of the modularity clustering problem for six graphs that are publicly available.
The sources of these six graphs as well as additional instances of the modularity clustering problem can be found in~\cite{cafieri2011locally}.
The six instances contain $34-115$ nodes and $561-6555$.

\subsubsection[Bird sound clustering]{Bird sound clustering\footnote{\url{https://keeper.mpdl.mpg.de/f/4803b50f0ab14e75b01c/?dl=1}}}
A Siamese network is trained by~\cite{stein2023correlation} to estimate wether the same species of bird can be heard in two different sound recordings.
Based on these estimates a multicut problem on the complete graph is solved to cluster sound recordings.
The names of the instances in this dataset indicate different settings: ``2s''/``3s'' indicates whether the data points are 2 or 3 second chunks of bird sound recordings; ``small''/``big'' indicates whether the dataset with 23 or 68 species is used; ``with-augments''/``no-augments'' indicates whether data augmentation is used for training; the rest of the file name indicates the underlying dataset as described in~\cite{stein2023correlation}.
In total there are $30$ instance with $3326-22252$ nodes and $5529475-247564626$ edges.

\subsection{Algorithms}
\begin{description}
\item[CGC~\cite{beier2014cut}:] Cut, Glue \& Cut, a heuristic that alternatingly bipartitions the graph and exchanges nodes in pairs of clusters via max-cut computed by a reduction to perfect matching.
\item[Fusion Moves~\cite{beier2017multicut}:] Explore subspaces of multicuts generated randomly via solving it with ILP-solvers.
\item[GAEC \& KLj~\cite{keuper2015efficient}:] The greedy additive edge contraction (GAEC) iteratively takes the most attractive edge and contracts it until no attractive edge is present.
Kernighan \& Lin with joins (KLj) computes sequences of node exchanges between clusters and cluster joins that improve the objective.
\item[Balanced Edge Contraction~\cite{kardoost2018solving}:] Like GAEC but additionally preferring edges with endpoints that contain fewer original nodes.
\item[Greedy Edge Fixation~\cite{levinkov2019comparative}:] Like GAEC but additionally prevent contractions on certain repulsive edges.
\item[Multi-stage Multicuts (MSM)~\cite{ho2021msm}:] Solve multiple minimum cost multicut problems across distributed compute units.
\item[Benders decomposition~\cite{lukasik2020benders}:] A Benders decomposition algorithm with node subproblems solved in parallel and accelerated through Magnanti-Wong Benders rows.
\item[RAMA~\cite{abbas2021combinatorial}:] Primal/dual algorithm using parallel edge contraction and fast separation/message passing for computing lower bounds and reduced costs.
\item[Message Passing~\cite{swoboda2017message}:] Sequential separation/message passing.
\item[Cycle Packing \& Persistency~\cite{lange2018partial}:] Fast cycle separation with interleaved greedy cycle packing for obtaining dual lower bounds. Simple persistency criteria for fixing variables to optimal values without optimizing the whole problem.
\item[Combinatorial Persistency Criteria~\cite{lange2019combinatorial}:] A number of persistency criteria for fixing variable to their optimal values ranging from easy enumerative to more involved ones which require optimization of larger subproblems.
\item[LP-rounding~\cite{demaine2006correlation}:]
    Solving an LP-relaxation and subsequent region growing rounding with approximation guarantees for the correlation clustering formulation.
\end{description}

\clearpage
\section{Asymmetric Multiway Cut}
The asymmetric multiway cut proposed in~\cite{kroeger2014asymmetric} is an extension of the multiway cut problem that allows for intra-class splits.

\begin{definition}[Asymmetric Multiway Cut]
Given a weighted graph $G=(V,E,c_E)$, $K$ classes, $P \subset [K]$ partitionable classes and node costs $c_V : [K]^V \rightarrow \R$ the asymmetric multiway cut problem is
\begin{equation}
    \begin{array}{rl}
    \min\limits_{\substack{x \in V \rightarrow [K]\\ y \in \mathcal{M}_G}}
    & \sum_{i \in V} c_V(i,x_i) + \sum_{ij \in E} c_E(e) \cdot y_e \\
    \text{s.t.}
    & x_i \neq x_j \Rightarrow y_e = 1\ \forall ij \in E \\
    & x_i = x_j \ \&\ x_i \notin P \Rightarrow y_e = 0\ \forall ij \in E \\
    \end{array}
\end{equation}
\end{definition}

\begin{figure}[H]
    \begin{center}
        \includegraphics[width=0.5\columnwidth]{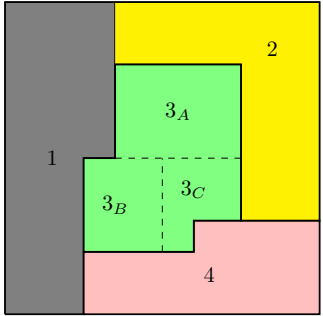}
    \end{center}
    \caption{Illustration of an asymmetric multiway cut with 4 classes and with class 3 having internal boundaries producing three clusters $3_A$, $3_B$ and $3_C$.}
    \label{fig:asymmetric-multiway-cut}
\end{figure}

An illustration of asymmetric multiway cut is given in Figure~\ref{fig:asymmetric-multiway-cut}.

\subsection{File Format}

\begin{fileformat}
ASYMMETRIC MULTIWAY CUT
PARTITIONABLE CLASSES
(*$p_1$*) ... (*$p_{\abs{P}}$*)

NODE COSTS
(*$c_1(1)$*) ... (*$c_1(K)$*) 
.
.
.
(*$c_{\abs{V}}(1)$*) ... (*$c_{\abs{V}}(K)$*)  

EDGE COSTS
(*$i_1$*) (*$j_1$*) (*$c_1$*)
.
.
.
(*$i_{\abs{E}}$*) (*$j_{\abs{E}}$*) (*$c_{\abs{E}}$*)
\end{fileformat}
where $P = \{p_1,\ldots,p_{\abs{P}}\}$ and
$E = \{i_1 j_1,\ldots,i_{\abs{E}} j_{\abs{E}}\}$.

\subsection{Datasets}

\subsubsection{Panoptic Segmentation}
In~\cite{abbas2021combinatorial} the asymmetric multiway cut solver was used for panoptic segmentation on the following two datasets.

\paragraph{Cityscapes}
Contains traffic related images from~\cite{cordts2016cityscapes} of resolution $1024 \times 2048$\footnote{\url{https://keeper.mpdl.mpg.de/f/63bfbea779d042a59a4d/?dl=1}} and $4 \times$ downsampled ones at resolution $256 \times 512$\footnote{\url{https://keeper.mpdl.mpg.de/f/956f3e74320a432295d9/?dl=1}}.
Images contain $21$ classes divided into $8$ \lq thing\rq\ and $11$ \lq stuff\rq\ classes.
$100$ easy and $100$ hard to segment images from the validation set were chosen for the benchmark. 

\paragraph{COCO}
Contains diverse images from~\cite{lin2014microsoft} of resolution $640 \times 480$\footnote{\url{https://keeper.mpdl.mpg.de/f/25852c04694047ce8e70/?dl=1}} and $4 \times$ downsampled to $160 \times 120$\footnote{\url{https://keeper.mpdl.mpg.de/f/dcdb2b8dc2f34b21a9bf/?dl=1}} with $133$ classes divided into $80$ \lq thing\rq\ and $53$ \lq stuff\rq\ classes.
$100$ easy and $100$ hard to segment images from the validation set were chosen for the benchmark. 

\subsection{Algorithms}
\begin{description}
\item[ILP~\cite{kroeger2014asymmetric}:] ILP description and cutting plane procedures for separating the defining inequalities.
\item[GAEC for AMWC~\cite{abbas2021combinatorial}:] Generalization of the multicut solver GAEC~\cite{keuper2015efficient} for asymmetric multiway cut.
\end{description}

\clearpage
\section{Factorized Complete Multicut}
\label{sec:complete_multicut}
Factorized Complete Multicut is an extension of the multicut problem to complete graphs where the edge costs are computed from node features.
The problem was proposed in~\cite{aabbas23_clusterfug} to alleviate the need for specifying a suitable graph structure for the multicut problem~\eqref{eq:multicut}. 
\begin{definition}[Factorized Complete Multicut]
    Let a set of nodes $V$ where each node $i \in V$ with feature vectors $f_i \in \R^d$ be given.
    We consider all possible edges i.e.\ $E = \{ij : i \in V, j \in V\setminus \{i\} \}$. 
    Let $G = (V, E)$ be the complete graph and $s: \R^d \times \R^d \rightarrow \R$ the edge cost function.
    The factorized complete multicut problem is
    \begin{equation}
        \label{eq:factorized_complete_multicut}
        \begin{array}{rl}
            \min_{y \in \mathcal{M}_G} \sum\limits_{i \in V} \sum\limits_{j \in V \setminus i} s(f_i, f_j) \cdot y_{ij}.
        \end{array}
    \end{equation}
\end{definition}

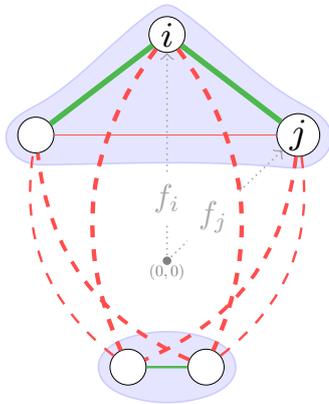
\begin{figure}[H]
    \begin{center}
    \scalebox{1.5}{\def\connect#1#2#3{%
    \path let
      \p1 = ($(#2)-(#1)$),
      \n1 = {veclen(\p1)},
      \n2 = {atan2(\y1,\x1)} 
    in
      (#1) -- (#2) node[#3, midway, rotate = \n2, shading angle = \n2+90, minimum        width=\n1, minimum height=1pt, inner sep=1pt] {};
}
\begin{tikzpicture}
\definecolor{scattercolor1}{HTML}{800080}
\definecolor{scattercolor2}{HTML}{DD0000}
\definecolor{scattercolor3}{HTML}{009900}
\definecolor{scattercolor4}{HTML}{808000}
\definecolor{scattercolor5}{HTML}{D30E78}
\definecolor{scattercolor6}{HTML}{036ffc}
\definecolor{scattercolor7}{HTML}{eb9800}
\tikzstyle{cut-edge}=[gray!70, dashed]
\tikzstyle{vector-arrow}=[gray!70, densely dotted]
\tikzstyle{att-edge}=[scattercolor3!70]
\tikzstyle{rep-edge}=[red!70]
\tikzstyle{vertex}=[circle, draw, ultra thin, fill=white, inner sep=0.7pt, minimum width=9pt]
\tikzstyle{vector}=[rectangle, draw,inner sep=1pt,outer sep=0pt,
minimum height=5mm, minimum width=2mm, ultra thin]

\node[vertex] (Pi) at ({90}:2.0cm) {\small{$i$}};
\node[vertex] (Pj) at ({44}:1.6cm) {\small{$j$}};
\node[vertex] (Pk) at ({290}:1.cm) {};
\node[vertex] (Pl) at ({250}:1.cm) {};
\node[vertex] (Pm) at ({136}:1.6cm) {};

\draw[->, style=vector-arrow] (0,0) -- (Pi) node [pos=0.3,fill=white,scale=0.8] {{$f_i$}};
\draw[->, style=vector-arrow] (0,0) -- (Pj) node [pos=0.4,fill=white,scale=0.8] {{$f_j$}};

\draw[fill=gray, gray] (0,0) circle (.2ex);
\node[draw=none, gray, scale=0.4] (origin) at (-0.0, -0.1) {$(0, 0)$};
\draw[style=att-edge, line width = 1.4pt] (Pi) -- (Pj);
\draw[style=rep-edge, line width = 1.1pt, dashed] (Pi) to [bend left=30] (Pk);
\draw[style=rep-edge, line width = 1.1pt, dashed] (Pi) to [bend right=30] (Pl);
\draw[style=att-edge, line width = 1.4pt] (Pi) -- (Pm);
\draw[style=rep-edge, line width = 0.6pt, dashed] (Pj) to [bend left=30] (Pk);
\draw[style=rep-edge, line width = 0.95pt, dashed] (Pj) to [bend left=30] (Pl); 
\draw[style=rep-edge, line width = 0.2pt] (Pj) -- (Pm); 
\draw[style=att-edge, line width = 0.6pt] (Pk) -- (Pl); 
\draw[style=rep-edge, line width = 0.95pt, dashed] (Pk) to [bend left=30] (Pm); 
\draw[style=rep-edge, line width = 0.6pt, dashed] (Pl) to [bend left=30] (Pm); 

\begin{pgfonlayer}{background}

\draw[fill=blue!50,opacity=0.2,draw=blue]([yshift=0.1cm] Pi.north) to[closed,curve through={([yshift=0.1cm]Pi.north east) .. ([yshift=0.1cm]Pj.north) .. ([xshift=0.1cm]Pj.east) .. ([yshift=-0.1cm]Pj.south) .. ([yshift=-0.15cm]Pj.south west) .. ([yshift=-0.15cm]Pm.south east) .. ([yshift=-0.1cm]Pm.south) .. ([xshift=-0.1cm]Pm.west) .. ([yshift=0.1cm]Pm.north) .. ([yshift=0.1cm]Pi.north west)}]([yshift=0.1cm] Pi.north);

\draw[fill=blue!50,opacity=0.2,draw=blue]([yshift=0.1cm] Pk.north) to[closed,curve through={([yshift=0.1cm]Pl.north) .. ([xshift=-0.1cm]Pl.west) ([yshift=-0.1cm] Pl.south) .. ([yshift=-0.1cm]Pk.south)}]([xshift=0.1cm] Pk.east);

\end{pgfonlayer}

\end{tikzpicture}}
    \end{center}
    \caption{Illustration of factorized complete multicut problem on $5$ nodes. Each node $i$ is associated with a vector $f_i \in \R^2$ and all possible edges between distinct nodes are considered. The edge cost between a pair of nodes $i$, $j$ is given by $s(f_i, f_j)$ and attractive/repulsive edges are colored green/red. Edge thickness represents absolute edge cost. Also shown is the optimal partitioning to $2$ clusters with cut edges as dashed lines.}    
    \label{fig:factorized-complete-multicut}
\end{figure}

An illustration of factorized complete multicut is given in Figure~\ref{fig:factorized-complete-multicut}.

In the following we assume that the edge cost $s(f_i, f_j)$ is defined as $\la f_i, f_j \ra - \alpha$ as done in~\cite{aabbas23_clusterfug}. Increasing value of $\alpha$ helps in creating more clusters and vice versa.
\subsection{File Format}
\begin{fileformat}
FACTORIZED COMPLETE MULTICUT
(*$\alpha$*)
(*$\abs{V}$*) (*d*)
(*$f_1^1$*) ... (*$f_1^d$*)
.
.
(*$f_{\abs{V}}^1$*) ... (*$f_{\abs{V}}^d$*)
\end{fileformat}
\subsection{Benchmark datasets}

\subsubsection{Imagenet Clustering}
consists of the ImageNet~\cite{deng2009imagenet} validation set containing $50k$ images.
Each image acts a node for the complete multicut problem.
The features have a dimension of $2048$ and are normalized to have unit $L_2$ norm.
Two problem instances created by~\cite{aabbas23_clusterfug} are provided\footnote{\url{https://keeper.mpdl.mpg.de/f/ac94cee3de7b45ed9539/?dl=1}} containing $5k$ and $50k$ images by considering $100$ and all $1000$ classes respectively. 
All instances use $\alpha = 0.4$.

\subsubsection{Cityscapes Instance Segmentation}
for instance segmentation on the Cityscapes validation set~\cite{cordts2016cityscapes}.
For each \textit{thing} class a separate complete multicut problem instance is created at $4$ times downsampled resolution w.r.t.\ the input resolution. 
In total there are $1631$ problem instances\footnote{\url{https://keeper.mpdl.mpg.de/f/e5c3a48377a34561991d/?dl=1}} from~\cite{aabbas23_clusterfug}.
The largest problem instance contains around $43k$ nodes. All instances use $\alpha = 0.4$.
\subsection{Algorithms}
For solving the complete multicut problem all algorithms from Section~\ref{sec:multicut} are applicable but can be inefficient for large problems. 
\begin{description}
\item[ClusterFuG~\cite{aabbas23_clusterfug}:] Multiple algorithms based on edge contraction similar to GAEC algorithm for multicut~\cite{keuper2015efficient}.
    The algorithms make use of nearest neighbour techniques and factorized edge costs structure for efficiency.
\end{description}

\clearpage
\section{Lifted Disjoint Paths}
\label{sec:lifted-disjoint-paths}

The lifted disjoint paths problem is an extension of the node-disjoint paths problem by additional lifted edges that represent connectivity given by the disjoint paths.
They allow to express long-range connectivity priors.

\begin{definition}[Flow Network and Lifted Graph]
Consider two directed acyclic graphs $G = (V, E)$ and $G' = (V',E')$ where $V' =V\backslash \{s,t\}$.
The graph $G=(V,E)$ represents the flow network and we denote by $G'$ the lifted graph.
The two special nodes $s$ and $t$ of $G$ denote source and sink node respectively.
\end{definition}

\begin{definition}[Paths]
We define the set of paths starting at $v$ and ending in $w$ as
\begin{equation}
vw\mhyphen\text{paths}(G) = \{
(v_1v_2,\ldots,v_{l-1}v_l) : \begin{array}{c} v_i v_{i+1} \in E \\ v_1 = v, v_l = w \end{array}
 \}
\end{equation}
For a $vw\mhyphen\text{path}$ $P$ we denote its edge set as $P_E$ and its node
set as $P_V$. 
\end{definition}

\begin{definition}[Lifted Edges]
Given an indicator vector $y \in \{0,1\}^E$ of a node-disjoint set of paths and a lifted edge $e \in E'$ its indicator variable $y'_e \in \{0,1\}$ is defined as
\begin{equation}
\label{eq:lifted-disjoint-paths-lifted-edge}
y'_e \Leftrightarrow \exists P \in vw\mhyphen\text{paths}(G) \text{ s.t. } \forall ij \in P_E : y_{ij} = 1\,.
\end{equation}
\end{definition}

\begin{definition}[Lifted Disjoint Paths Problem]
Given edge costs $c \in \R^E$, node cost $\omega \in \R^V$ for the flow network $G$ and edge costs $c' \in \R^{E'}$ for the lifted graph $G'$ we define the lifted disjoint paths problem as
\begin{equation}
\label{eq:lifted-disjoint-paths}
\begin{array}{cl}
\min\limits_{\substack{y \in \{0,1\}^E, y' \in \{0,1\}^{E'} \\ x \in \{0,1\}^V }} 
&
\la c, y \ra + \la c', y' \ra + \la \omega, x \ra \\
\text{s.t.}
& y \text{ node-disjoint } s,t-\text{flow in } G \\
& x \text{ flow through nodes of } G \\
& y,y' \text{ feasible according to~\eqref{eq:lifted-disjoint-paths-lifted-edge}}\,.
\end{array}
\end{equation}
\end{definition}

\begin{figure}
	\centering
	\includegraphics[width=0.9\columnwidth]{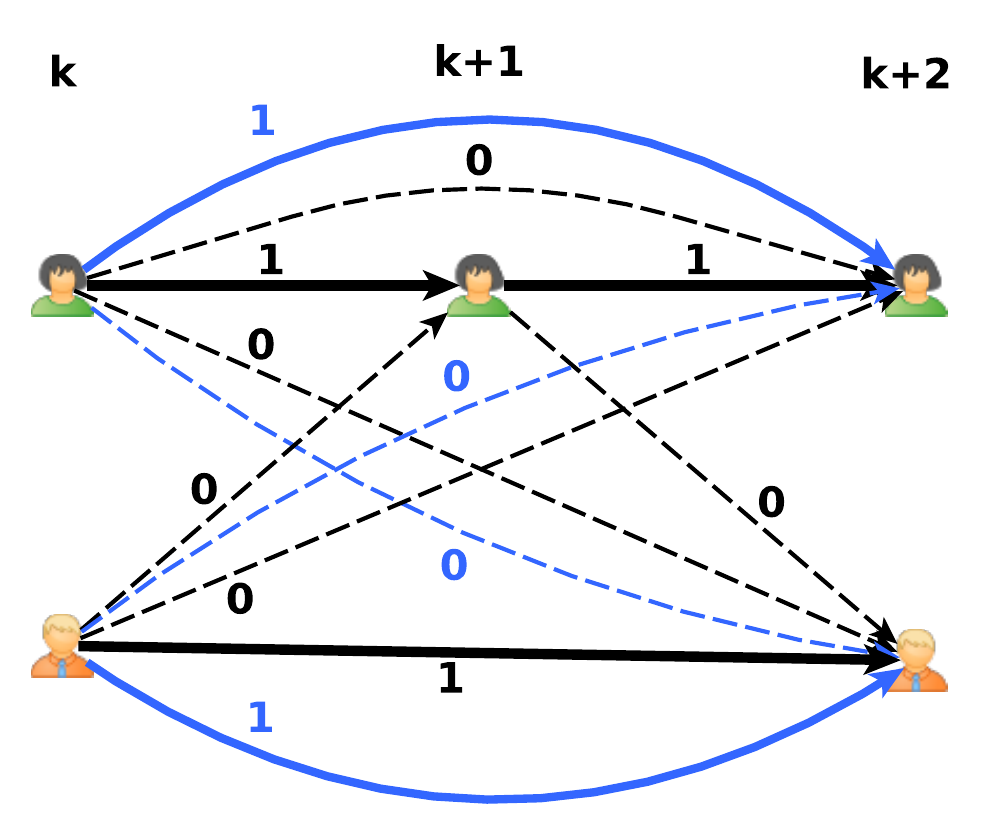}
	\caption{Illustration of a lifted disjoint paths problem on flow network (black edges) and lifted graph (blue edges).
		Active edges $y,y' = 1$ are solid while inactive edges $y,y' = 0$ are dashed.}
	\label{fig:LDP}	
\end{figure}

An ILP formulation of~\eqref{eq:lifted-disjoint-paths} is given in~\cite{hornakova2020lifted}.
For an illustration of the lifted disjoint paths problem see Figure~\ref{fig:LDP}.


\subsection{File Format}
\paragraph{File defining costs between pairs of nodes.} Some of the pairs are selected to create flow edges, some of them are selected to create lifted edges according to the strategy for graph creation given by solver parameters. Note that lifted and flow edges can overlap. 
\begin{fileformat}
(*$|V'|$*)
(*$v_1$*),(*$w_1$*),(*$c_1$*)
.
.
.
(*$v_m$*),(*$w_m$*),(*$c_m$*) 
\end{fileformat}
\paragraph{File defining the distribution of nodes in time frames.} Time frames are numbered from $1$ to $T$ (maximal time frame). Nodes are numbered from $0$ to $|V'|-1$. $V_i$ denotes nodes in time frame $i$.
\begin{fileformat}
(*$1$*) (*$0$*)
(*$1$*) (*$1$*)
.
.
.
(*$1$*) (*$|V_1|-1$*)
(*$2$*) (*$|V_1|$*)
.
.
.
(*$2$*) (*$|V_1|+|V_2|-1$*)
.
.
.
(*$T$*) (*$|V'|-1$*)
\end{fileformat}
\paragraph{File with solver parameters.}The file starts with the keyword [SOLVER] (including the brackets). Each parameter name is written with capital letters and underscores. Parameter values follow the $=$ sign.
\begin{fileformat}
[SOLVER]
FIRST_PARAMETER_NAME=value1
SECOND_PARAMETER_NAME=value2
.
.
.
\end{fileformat}

\paragraph{Input file for the approximate solver.} We typically use the same set of parameters for processing all instances within a~dataset. The first LDP solver LifT specifies path to instance files in the parameter file. Therefore, a~separate parameter file is needed for each problem instance. Solver ApLift enables to use one parameter file for multiple instances. The path to instance files and to the parameter file is specified in the input input file that has the following format. 
\begin{fileformat}
INPUT_GRAPH=/path/to/edge/costs
INPUT_FRAMES=/path/to/nodes/in/frames
INPUT_PARAMS=/path/to/parameter/file
\end{fileformat}

\subsection{Datasets}
\subsubsection{MOT Challenge}
The main application of LDP is in multiple object tracking.
The MOT15/16/17 benchmarks \cite{MOTChallenge2015,MOT16} contain semi-crowded videos sequences filmed from a~static or a~moving camera. 
MOT20 \cite{MOTChallenge20} comprises crowded scenes with considerably higher number of  frames  and detections per frame.

\subsection{Algorithms}
\begin{description}
    \item[ILP Solver~\cite{hornakova2020lifted}:] Cutting planes for ILP solvers together with a two-stage procedure for computing tracklets first.
    \item[Message Passing Solver~\cite{hornakova2021making}:] Decompose the full problem into combinatorially solved subproblems and link them together with Lagrange multipliers updated by block coordinate ascent.
\end{description}

\clearpage
\section{Graph Matching}
\label{sec:graph-matching}

The graph matching problem is equivalent to Lawler's form of the quadratic assignment problem (QAP) from the combinatorial optimization literature.

Given two point sets the task is to establish partial one-to-one correspondences between points in the respective sets.

\begin{definition}[Graph Matching]
Given $N_1,N_2 > 0$ and linear costs $c_{ij} \in \R$ for $(i,j) \in \R^{N_1 \times N_2}$ and quadratic costs $d \in \R^{N_1 \times N_2 \times N_1 \times N_2}$ the graph matching problem is
\begin{equation}
    \begin{array}{rl}
    \min\limits_{x \in \R^{N_1 \times N_2}} & \sum\limits_{ij \in E} c_{ij} x_{ij} + \sum_{ijkl \in T} d_{ijkl} x_{ij} x_{kl} \\ 
    \text{s.t.} 
    & \sum\limits_{j=1,\ldots,N_1} x_{ij} \leq 1 \quad \forall i\in [N_1] \\
    & \sum\limits_{i=1,\ldots,N_2} x_{ij} \leq 1 \quad \forall j\in [N_2] \\
    \end{array}
\end{equation}
\end{definition}

\begin{figure}[H]
    \begin{center}
        \includegraphics[width=0.95\columnwidth]{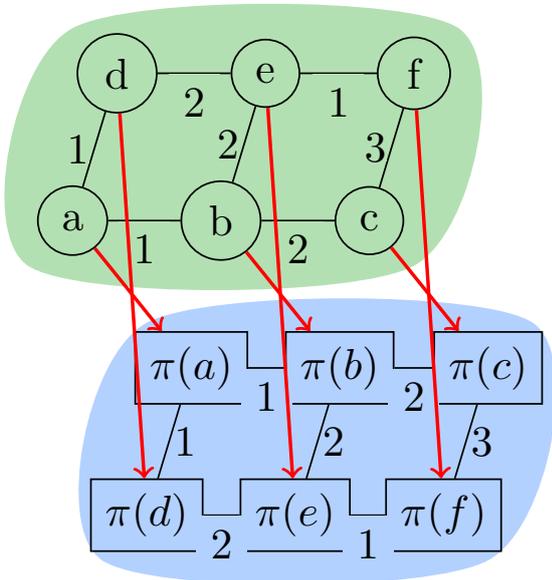}
        \caption{Finding node correspondence $\pi$ on two graphs (green and blue background respectively). Nodes are matched (red lines) such that matched edges have corresponding weights.}
        \label{fig:cell-tracking}
    \end{center}
\end{figure}

\subsection{File Format}
\label{sec:graph-matching-file-format}

We use the file format introduced by~\cite{torresani2012dual}.

\begin{fileformat}
p (*$N_0$*) (*$N_1$*) A E 
a 1 (*$i_1$*) (*$j_1$*) (*$c_1$*)
.
.
.
a A (*$i_A$*) (*$j_A$*) (*$c_A$*)
e (*$a_1$*) (*$a'_1$*) (*$d_1$*)
.
.
.
e (*$a_E$*) (*$a'_E$*) (*$d_E$*)
\end{fileformat}
where $A$ is the number of linear assignment terms (those that are not listed are assumed to have value $\infty$) and $E$ is the number of pairwise terms (those that are not listed are assumed to have value $0$).
Linear assignments start with the identifier $a$ and are numbered from $1$ to $A$. They go from node $i_l$ to $j_l$ with cost $c_l$ for $l \in [A]$.
Quadratic costs start with the identifier $e$ and have cost $d_l$, $l \in [E]$.
The quadratic cost is incurred if both the $a_l$-th and $a'_l$-th assignments are active.

We also provide LP files for all instances.

\subsection{Benchmark datasets}

\subsubsection[Hotel \& House]{Hotel \& House\footnote{\url{https://keeper.mpdl.mpg.de/f/0fe3f173da55491cb10a/?dl=1
}}}
Matching of keypoints in two rigid objects~\cite{torresani2012dual}.
Images of the same object are taken at different angles and correspondences are computed from costs derived by appearance and geometric terms.
There are 105 instances for each object with $N_1 = N_2 = 30$ and dense linear and quadratic assignments.



\subsubsection[Worms]{Worms\footnote{\url{https://keeper.mpdl.mpg.de/f/1058b4e0d8664c0bb0d5/?dl=1}}}
Matching nuclei of C.\ elegans for automatic annotation~\cite{kainmueller2014active} in bio-imaging. 
The 30 instances have up to $\max(N_1,N_2) = 1500$ points. Costs are sparse.

\subsection{Algorithms}
\begin{description}
\item[Semidefinite Programming~\cite{schellewald2005probabilistic}:]
    Semidefinite relaxation for graph matching.
\item[Dual Decomposition~\cite{torresani2012dual}:]
Subgradient ascent on a dual decomposition of graph matching into max-flow, linear assignment and enumerative local subproblems.
\item[Hungarian BP~\cite{zhang2016pairwise}:]
Message passing using additionally a linear assignment solver.
\item[Message Passing~\cite{swoboda2017study}]
Several improved message passing schemes on different decompositions.
\item[Fusion Moves~\cite{hutschenreiter2021fusion}:] Explore subspaces of graph matchings generated randomly and solve with ILP solvers.
\item[Covering Trees~\cite{yarkony2010covering}:] Convert the graph matching problem to one large tree MRF with additional Lagrange multipliers and optimize with message passing.
\item[DS*~\cite{bernard2018ds}:]
    A lifting-free convex relaxation approach.
\item[Semidefinite Optimization~\cite{kezurer2015tight}:]
    Semidefinite optimization problem formulation of multi-graph matching.
\item[Graduated assignment~\cite{gold1996graduated}:]
    An algorithm combining graduated nonconvexity, two-way assignment constraints and sparsity.
\item[Tabu search~\cite{adamczewski2015discrete}:]
    Reformulation of graph matching into an equivalent weighted maximum clique problem solved by tabu search.
\item[Fixed-point iteration~\cite{leordeanu2009integer}:]
    Climbing and convergence guaranteeing discretization scheme.
\item[Max-Pooling~\cite{cho2014finding}:]
    Matching through evaluating neighborhood candidates and gradually propagating matching scores to neighbors via max-pooling, eventually producing reliability matching scores.
\item[Spectral correspospondence finding~\cite{leordeanu2005spectral}:]
    Correct assignment clustering via principal eigenvector and additionally imposing mapping constraints.
\item[Umeyama's Eigen-decomposition method~\cite{zhao2007using}:]
    Generalization of Umeyama's formula for graph matching and corresponding matching algorithm.
\item[Reweighted random walks~\cite{cho2010reweighted}:]
    Graph matching formulation as node selection on an association graph solved by simulating random walks with reweighting jumps enforcing matching constraints.
\item[MCMC~\cite{lee2010graph}:]
    Data-driven Markov Chain Monte Carlo sampling additionally using spectral properties of the graphs.
\item[Sequential Monte Carlo~\cite{suh2012graph}:]
    Sequential Monte Carlo sampling a sequence of of intermediate target distributions with importance resampling for maximizing the graph matching objective.
\item[Path following~\cite{zaslavskiy2008path}:]
    A convex-concave programming formulation obtained by rewriting graph matching as a least-squares problem on the set of permutation matrices and solving a quadratic convex and concave problem.
\item[Factorized graph matching~\cite{zhou2015factorized}:]
Graph matching via affinity matrix factorization as a Kronecker product of smaller matrices solved approximately through a path-following Frank-Wolfe algorithm.
\end{description}

\clearpage
\section{Multi-Graph Matching}
\label{sec:multi-graph-matching}
The multi-graph matching problem is an extension of the graph matching problem from Section~\ref{sec:graph-matching} to more than two graphs with additional cycle consistency constraints ensuring that compositions of matchings are consistent.

\begin{definition}[Multi-Graph Matching]
Given $N_1,\ldots,N_K > 0$ and linear costs $c^{pk} \in \R^{N_p \times N_k}$ and quadratic costs $d^{pk} \in \R^{N_p \times N_k \times N_p \times N_k}$ for every $p < k$, the multi-graph matching problem is
\begin{equation}
\begin{array}{cl}
    \min\limits_{(x^{pk} \in \R^{N_p \times N_k})_{p,k}}
    &\sum\limits_{p < k} \sum\limits_{ij \in E} c_{ij} x_{ij} + \sum\limits_{ijkl \in T} d_{ijkl} x_{ij} x_{kl} \\ 
\text{s.t.}
& \sum\limits_{j=1,\ldots,m} x^{pk}_{ij} \leq 1 \quad \forall i\in [N_p], p < k \\
& \sum\limits_{i=1,\ldots,n} x^{pk}_{ij} \leq 1 \quad \forall j\in [N_k], p < k \\
& x^{pk} \cdot x^{kl} \leq x^{pl} \quad \forall p \neq k \neq l \\
    & x^{pk} = (x^{kp})^\top \quad \forall p \neq k
\end{array}
\end{equation}
where we use~$\cdot$~for matrix multiplication and~$\leq$~holds elementwise.
\end{definition}

\begin{figure}[H]
    \centering
    \includegraphics[width=0.7\columnwidth]{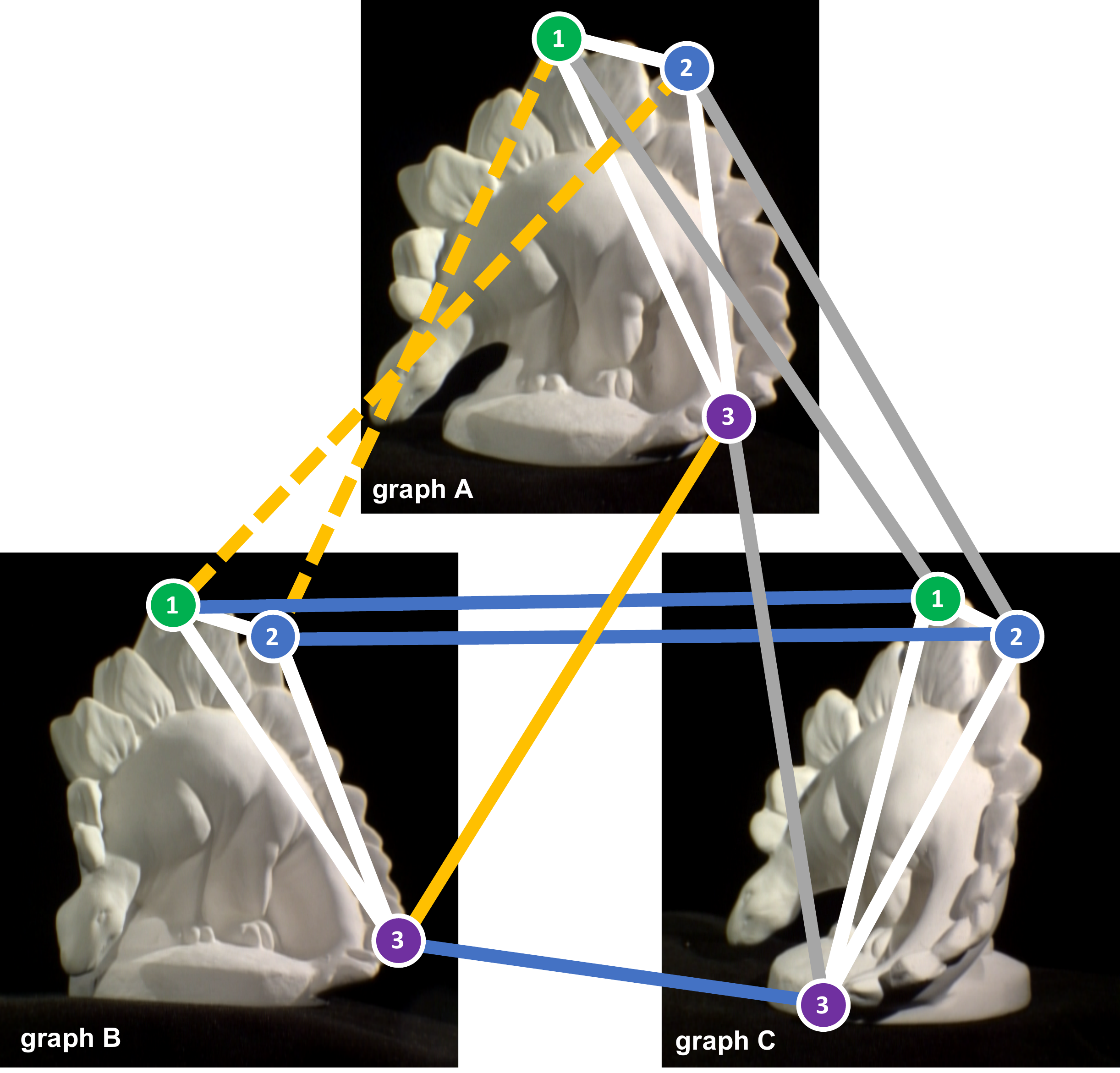}
    \caption{
        Illustration of cycle consistency in multi-graph matching (best viewed in color).
        Each graph $A$, $B$, $C$ comprises three nodes (green, blue, purple) and three edges (white lines).
        The true correspondence is indicated by the node colour and node labels 1, 2, 3.
        Matchings between pairs of graphs are shown by coloured lines ($A\leftrightarrow B)$ in yellow, $A \leftrightarrow C$ in gray, and $B \leftrightarrow C$ in blue).
        Wrong matchings are indicated by dashed lines.
        The multi-matching $A1 \leftrightarrow B2 \leftrightarrow C2 \leftrightarrow A2$ is not cycle consistent.
    }
    \label{fig:multi-graph-matching}
\end{figure}

An illustration of the multi-graph matching problem with matchings that obey resp.\ violate cycle consistency is given in Figure~\ref{fig:multi-graph-matching}.

\subsection{File Format}
The file format is derived from the one for graph matching described in Section~\ref{sec:graph-matching-file-format}.

\begin{fileformat}
gm (*$0$*) (*$1$*)
### graph matching problem for ###
### matching (*$0$*) with (*$1$*) ###
### as in Section(*$\text{~\ref{sec:graph-matching-file-format}}$*) ###

.
.
.

gm (*$K-2$*) (*$K-1$*)
### graph matching problem for ###
### matching (*$K-1$*) with (*$K-2$*) ###
### as in Section(*$\text{~\ref{sec:graph-matching-file-format}}$*) ###
\end{fileformat}

\subsection{Datasets}

\subsubsection[Hotel \& House]{Hotel \& House\footnote{\url{https://keeper.mpdl.mpg.de/f/d7ba7019eede485fb457/?dl=1}}}
In this experiment we consider the CMU house and hotel sequences.
$40\%$ of the points are outliers and the total number of points per image is $10$. 
For problem generation we have followed the protocol of [41], where further details are described.
Costs are computed as in~\cite{yan2015multi}.
In total there are 80 instances, 40 for house and 40 for hotel.

\subsubsection[Synthetic]{Synthetic\footnote{\url{https://keeper.mpdl.mpg.de/f/8ca51384836d40a09487/?dl=1}}}
Four different categories (complete, density, deform, outlier) of synthetic multi-graph matching problems with the number of point sets varying from 4 to 16 generated as in~\cite{yan2015multi}.
In total there are 160 instances, 40 from each category.

\subsubsection[Worms]{Worms\footnote{\url{https://keeper.mpdl.mpg.de/f/8b1385731fae4febb4fe/?dl=1}}}
Multi-graph matching instances generated from 30 images of C.\ elegans~\cite{kainmueller2014active}.
Points correspond to nuclei of the organism, which can number up to 558 per point set.
Instances have from 3 to 10 randomly selected point sets.
In total there are 400 instances, 50 per number of point sets.

\subsection{Algorithms}
\begin{description}[style=unboxed]
    \item[Permutation Synchronization~\cite{bernard2019synchronisation,pachauri2013solving}:]
        Non-negative matrix factorisation followed by Euclidean projection for binarization.
    \item[Alternating Graph matching~\cite{yan2013joint,zhou2015multi,yan2015consistency}:]
        Alternating optimization between graph matching solvers and a cycle consistency-enforcing component.
    \item[Graduated Consistency~\cite{yan2014graduated,yan2015multi}:]
        Iterative approximation of the graph matching objective with gradual consistency enforcement.
    \item[Matrix Decomposition~\cite{yan2015matrix}:]
         Matrix decomposition based formulation solved through convex optimization.
    \item[Factorized Matching~\cite{zhou2015factorized}:]
        Global alternating minimization approach via low-rank matrix recovery.
    \item[Semidefinite Optimization~\cite{kezurer2015tight}:]
        Semidefinite optimization problem formulation of multi-graph matching.
    \item[Fast Clustering~\cite{tron2017fast}:]
        Clustering-based formulation identifying multi-image matchings from a density function in feature space.
    \item[Tensor Power Iteration~\cite{shi2016tensor}:]
        Rank-1 tensor approximation solved via power iteration.
    \item[Mining Consistent Features~\cite{wang2018multi}:]
        Multi-graph matching by matching sparse feature sets and geometric consistency through low-rank constraints. 
    \item[DS*~\cite{bernard2018ds}:]
        A lifting-free convex relaxation approach.
    \item[Random Walk~\cite{park2019consistent}:]
        A multi-layer random walk synchronization approach.
    \item[Convex Message Passing~\cite{swoboda2019convex}:] Lagrange decomposition optimized with message passing. Cycle consistency is enforced via a cutting plane procedure that adds tightening subproblems.
    \item[HiPPi~\cite{bernard2019hippi}:]
        A higher-order projected power iteration method. 
\end{description}

\clearpage
\section{Cell Tracking}
The cell tracking problem is to construct a lineage tree of cells across a time sequence with potentially dividing cells, see Figure~\ref{fig:cell-tracking} for an illustration.
We use a general formulation incorporating mutually exclusive cell candidates originally proposed in~\cite{funke2012efficient}.

\begin{figure}[H]
    \begin{center}
        \includegraphics[width=\columnwidth]{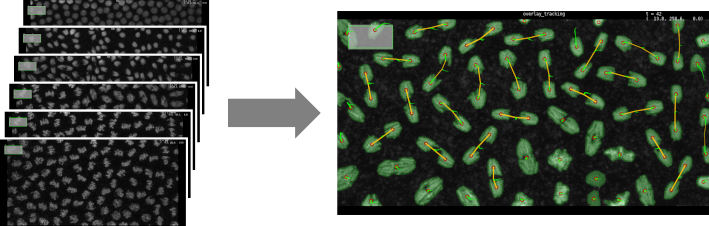}
        \caption{Cell tracking across timeframes.}
        \label{fig:cell-tracking}
    \end{center}
\end{figure}

\begin{definition}[Cell Tracking]
Let 
$T > 0$ time steps, 
detection candidates $V = V^1 \dot{\cup} \ldots \dot{\cup} V^T$ for each time step,
transition edges $M^t \subset V^t \times V^{t+1}$ and 
division edges $D^t \subset V^t \times \begin{pmatrix}V^{t+1} \\ 2 \end{pmatrix}$ for $t=1,\ldots,T-1$ and 
exclusion sets $\mathcal{E}^t \subset 2^{V^t}$ for $i=1,\ldots,T$
be given.
Define variables
\begin{equation}
\begin{array}{lrl}
    \text{Detection:} &
    x^{det,t} \in \{0,1\}^{V^t} & \forall t \in [T] \\
    \text{Appearance:} &
    x^{app,t} \in \{0,1\}^{V^t} & \forall t \in [T] \\
    \text{Disappearance:} &
    x^{disapp,t} \in \{0,1\}^{V^t} & \forall t \in [T] \\
    \text{Movement:} &
    y^{mov,t} \in \{0,1\}^{M^t} & \forall t \in [T-1] \\
    \text{Division:} &
    y^{div,t} \in \{0,1\}^{D^t} & \forall t \in [T-1] \\
   \end{array}
\end{equation}

The cell tracking problem is
\begin{multline}
\label{eq:cell-tracking-objective}
    \min_{\substack{x^{det}, x^{app}, x^{disapp}\\ y^{mov}, y^{div}}}  
    \la c^{det}, x^{det} \ra 
    + \la c^{app}, x^{app} \ra \\
    + \la c^{disapp}, x^{disapp} \ra 
    + \la c^{mov}, y^{mov} \ra 
    + \la c^{div}, y^{div} \ra 
\end{multline}
such that
    \begin{multline}
        \label{eq:cell-tracking-flow-conservation-incoming}
\forall t \in [T-1], i \in V^t: \\
        \sum_{j \in M^{t}} y^{mov,t}_{ij} + \sum_{j,k: ijk \in D^{t}} x^{div,t}_{ijk} = x^{det,t}_i + x^{disapp,t}_i 
    \end{multline}
    \begin{multline}
        \label{eq:cell-tracking-flow-conservation-outgoing}
\forall t \in [T-1], i \in V^t: \\
    \sum_{i \in M^{t}} y^{mov,t}_{ij} + \sum_{i,k: ijk \in D^{t}} x^{div,t}_{ijk} = x^{det,{t+1}}_j + x^{app,t}_i 
    \end{multline}
    \begin{multline}
        \label{eq:cell-tracking-exclusion-constraint}
        \forall Excl \in \mathcal{E}^t:\\
    \sum_{i \in Excl} x^{det,t}_i \leq 1 
\end{multline}
\end{definition}
The first two constraints in~\eqref{eq:cell-tracking-flow-conservation-incoming} and~\eqref{eq:cell-tracking-flow-conservation-outgoing} are flow conservation constraints stating that whenever a detection is active it must be linked to one detection in the previous timeframe and one or two in the next timeframe.
The third constraint~\eqref{eq:cell-tracking-exclusion-constraint} stipulates that for each exclusion set at most one detection might be active (e.g.\ when multiple cell candidates overlap spatially).

\subsection{File Format}
We use the file format also used in~\cite{haller2020primal}.

\begin{fileformat}
# comment line
# Cell detection hypotheses
H (*$t$*) (*$i$*) (*$c$*)
.
.
.

# Cell appearances, disappearances,
# movement and divisions
APP (*$t$*) (*$i$*) (*$c_{app}$*)
DISAPP (*$t$*) (*$i$*) (*$c_{app}$*)
MOVE (*$id$*) (*$i$*) (*$j$*) (*$c$*)
DIV (*$id$*) (*$i$*) (*$j$*) (*$k$*) (*$c$*)
.
.
.

# Exclusion constraints
CONFSET (*$i_1$*) + ... + (*$i_l$*) <= 1
.
.
.
\end{fileformat}

\begin{description}
    \item[\normalfont Comments] start with \#.
    \item[\normalfont Cell detections] start with `H', next comes the timeframe $t$, then the id $i$ and last the detection cost $c$. The ids must be unique also across timeframes.
    \item[\normalfont Cell appearance/disappearance] start with `APP' (resp.\ `DISAPP'), next comes the timeframe $t$, then the id $i$ and last the appearance/disappearance cost $c$.
    \item[\normalfont Cell movements] start with `MOVE', next come the movement id, then the ids of the two involved cell detections.
    \item[\normalfont Cell divisions] start with `DIV', next come the division id, then the ids of the three involved cell detections.
    \item[\normalfont Exclusion constraints] start with `CONFSET' and use the ids of the involved cell detections.
\end{description}

We provide LP files for all cell tracking instances as well.

\subsection{Datasets}

\subsubsection[AISTATS 2020 dataset]{AISTATS 2020 dataset\footnote{\url{https://keeper.mpdl.mpg.de/f/da232900c06c46399fd0/?dl=1
}}}
We have taken an extended set of instances used in~\cite{haller2020primal}.
The instances can be grouped as follows.

\paragraph{Drosophila embryo}
One problem instance for tracking nuclei in a developing Drosophila embryo.
The tracking models consists of up to 252 frames $\sim 320$ detection hypotheses each and $\sim 160$ true cells.

\paragraph{Flywing}
Tracking membrane-labelled cells in developing Drosophila flywing tissue.
Instances have up to 245 frames with $> 3300$ detection hypotheses per frame.

\paragraph{Cell Tracking Challenge (CTC)}
Publicly available cell tracking instances of sequences from~\cite{ulman2017objective}.
Instances have up to 426 frames and up to 1400 detection hypotheses.

\subsection{Algorithms}
\begin{description}
\item[Primal Dual Solver~\cite{haller2020primal}:] Use dual block coordinate ascent on a Lagrange decomposition and round primal solutions with a series of independent set problems.
\item[Generalized Network Flow~\cite{haubold2016generalized}:] Generalize the successive shortest path minimum cost flow algorithm to efficiently support cell divisions in a network flow model.
\end{description}

\clearpage
\section{Shape Matching}
\label{sec:shape-matching}
\begin{figure}[H]
	\begin{center}
		\includegraphics[width=0.8\columnwidth]{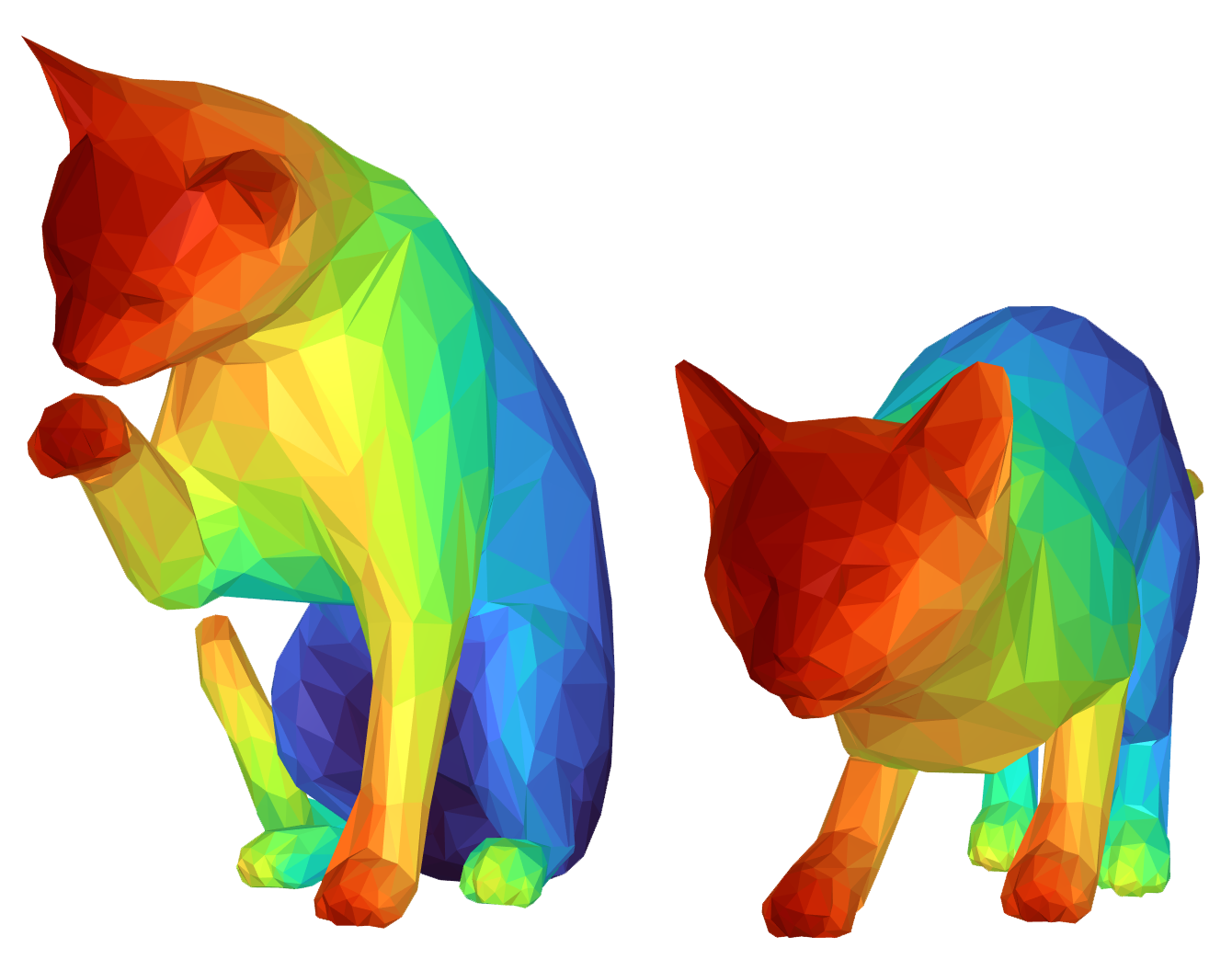}
	\end{center}
	\caption{Shape matching of a cat in different poses. Colors encode matches of triangles.}
	\label{fig:shape-matching}
\end{figure}

The shape matching problem is to find a mapping between shapes.
For an illustration see Figure~\ref{fig:shape-matching}.
We follow the shape matching formulation of~\cite{windheuser2011geometrically,windheuser2011large}. Here the shape matching problem is posed as an ILP which ensures an orientation preserving geometrically consistent matching.

\begin{definition}[Shape]
	We define a shape $X$ as a triplet $(V_X, E_X,F_X)$ of vertices $V_X$, 
	edges 
	$E_X \subset V_X \times V_X$ 
	and triangles 
	$F_X \subset V_X \times V_X \times V_X$,
	such that the manifold induced by the triangles is oriented and has no boundaries.
\end{definition}

\begin{definition}[Degenerate Triangles and Edges]
	By $\overline{F}_\bullet$ we denote the set of \emph{degenerate} triangles, that in addition to the triangles $F_\bullet$ also contains triangles formed by edges (a triangle with two vertices at the same position) and triangles formed by vertices (a triangle with three vertices at the same position).
	Similarly, we consider the set of degenerate edges $\overline{E}_\bullet$.
\end{definition}

\begin{definition}[Product Spaces]
	Let two shapes $X$ and $Y$ be given.
	The triangle product space is defined as
	\begin{equation}
		F \coloneqq \left\{ 
		\begin{pmatrix}
			a_1, b_1 \\
			a_2, b_2 \\ 
			a_3, b_3
		\end{pmatrix} \left|
		\begin{array}{ll}
			(a_1 a_2 a_3 \in F_X \wedge b_1 b_2 b_3 \in \overline{F}_Y )~\vee \\
			(a_1 a_2 a_3 \in \overline{F}_X \wedge b_1 b_2 b_3 \in {F}_Y ) 
		\end{array} \right.
		\right\}.\nonumber
	\end{equation}
	
	The edge product space is defined as
	\begin{equation}
		E \coloneqq \left\{ 
		\begin{pmatrix}
			a_1, b_1 \\
			a_2, b_2
		\end{pmatrix} \left|
		\begin{array}{ll}
			(a_1 a_2 \in E_X \wedge b_1 b_2 \in \overline{E}_Y )~\vee\\
			(a_1 a_2 \in \overline{E}_X \wedge b_1 b_2 \in {E}_Y )
		\end{array} \right.
		\right\}.\nonumber
	\end{equation}
\end{definition}

\begin{definition}[Discrete Surfaces]
	Let the triangle product space $F$ and the edge product space $E$ be given for a pair of shapes $X, Y$. A discrete surface $\Gamma \subset F$  is given by its indicator representation 
	\begin{equation}
		\Gamma = \{0, 1\}^{|F|}.
	\end{equation}
	The $f$-th entry $\Gamma_f$ belongs to the $f$-th triangle product in $F$.
\end{definition}
Triangle products encode a matching between a triangle from $X$ and a triangle from $Y$. Hence, discrete surfaces encode a matching between $X$ and $Y$. Constraints are needed to ensure surjectivity and geometric consistency of the matching. 

\begin{definition}[Projection Operator]
	The projection $\pi_X : F \rightarrow \Z^{\abs{F_X}}$ is defined by
	\begin{equation}
        (\pi_x \Gamma)_{a_1 a_2 a_3} = \sum_{\left\{ f \in F : f = \begin{pmatrix} a_1, b_1\\ a_2, b_2 \\ a_3, b_3 \end{pmatrix}\right\}} \Gamma_f\,.
	\end{equation}
    for all non-degenerate triangles $a_1 a_2 a_3 \in F_X$.

	The projection operator $\pi_Y$ is defined analoguously.
\end{definition}

\begin{definition}[Surjectivity Constraint]
	A given discrete surface $\Gamma$ ensures a surjective matching of $X$ with $Y$ if it projects to each shape $X$ and $Y$ such that
	\begin{equation}
			\begin{pmatrix}
			\pi_X \\ \pi_Y
		\end{pmatrix}
		\Gamma
		= 
		\begin{pmatrix}
			\boldsymbol{1}_{|F_X|} \\ \boldsymbol{1}_{|F_Y|}
		\end{pmatrix},
	\end{equation}
	with $\boldsymbol{1}_{|F_X|} \in \{1\}^{|F_X|}$ and $\boldsymbol{1}_{|F_Y|} \in \{1\}^{|F_Y|}$.
\end{definition}

\begin{definition}[Orientation]
	For the sets $E_X$ and $E_Y$ an arbitrary orientation shall be defined. This implies an orientation for each edge product $e = \begin{pmatrix} a_1,b_1 \\ a_2,b_2 \end{pmatrix} \in E$.
	By means of these orientations a vector $O\begin{pmatrix}e\end{pmatrix} \in \Z^{\abs{E}}$ can be defined for every edge product $e$. The entries of $O\begin{pmatrix}e\end{pmatrix}$ write as
	\begin{equation}
		O(e) = \begin{cases}
			1,& \emph{if } e \in E\\
			-1,& \emph{if } -e \in E\\
			0,& \text{otherwise}
		\end{cases}
	\end{equation}
\end{definition}

\begin{definition}[Boundary Operator]
	The boundary operator $\partial : F \rightarrow \Z^{\abs{E}}$ is defined for every triangle product by
	\begin{equation}
		\partial \begin{pmatrix} a_1, b_1 \\ a_2, b_2 \\ a_3, b_3 \end{pmatrix}
		=
		O\begin{pmatrix} a_1,b_1 \\ a_2,b_2 \end{pmatrix}
		+
		O\begin{pmatrix} a_2,b_2 \\ a_3,b_3 \end{pmatrix}
		+
		O\begin{pmatrix} a_3,b_3 \\ a_1,b_1 \end{pmatrix}
	\end{equation}
	where $a_i$ and $b_i$ form triangles in $X$ resp.\ $Y$ and $\begin{pmatrix} a_i,b_i \\ a_j,b_j \end{pmatrix}$ is the edge product connecting the product vertices $(a_i,b_i)$ and $(a_j,b_j)$.
\end{definition}

\begin{definition}[Closeness/Geometric Consistency Constraint]
	A given discrete surface $\Gamma$ ensures a geometric consistent matching of $X$ with $Y$, i.e. is a closed discrete surface if it satisfies
	\begin{equation}
		\partial
		\Gamma
		= 
		\boldsymbol{0}_{|E|}.
	\end{equation}
\end{definition}

\begin{definition}[Discrete Graph Surfaces]
	A graph surface is a discrete surface which fulfills the geometric consistency constraint as well as the surjectivity constraints.
\end{definition}

\begin{definition}[Deformation Energy]
	The deformation energy $ \mathbb{E}$ is defined for every triangle product. It describes the energy needed to deform the respective triangle from shape $X$ into the triangle from shape $Y$ and vice versa.
\end{definition}

We now have all operators to define the shape matching problem as the search for a discrete graph surface.
\begin{definition}[Shape Matching]
	Given an energy $\mathbb{E} \in \R^{\abs{F}}$ the shape matching problem is
	\begin{equation} 
		\underset{\Gamma \in \{0, 1\}^{|F|}}{\min} \mathbb{E}^\top  \Gamma  
		~~  \text{s.t.} ~~
		\begin{pmatrix}
			\pi_X \\ \pi_Y \\ \partial
		\end{pmatrix}
		\Gamma
		= 
		\begin{pmatrix}
			\boldsymbol{1}_{|F_X|} \\ \boldsymbol{1}_{|F_Y|} \\ \boldsymbol{0}_{\abs{E}} \\
		\end{pmatrix},
		\label{eq:shape-matching-opt}
	\end{equation}
\end{definition}

\subsection{File Format}
Instances are given in the ILP file format from Section~\ref{sec:ilp-file-format}. The filenames contain information about the respective problems
\begin{equation}
	\begin{aligned}
	&\texttt{<\# binvars>\_<\# constraints>\_<shape X>\_} \\
	&\texttt{<\# triangles>\_<shape Y>\_<\#  triangles>.lp}
	\end{aligned}
\end{equation}
Each binary variable belongs to a triangle product. Considering a vertices of shape $X$ as $a_i$ and a vertices of shape $Y$ as $b_i$ respectively. The names of the binary variables are structured as follows 
\begin{equation}
	\texttt{x} \_ \texttt{a}_1 \_ \texttt{a}_2 \_ \texttt{a}_3 \_\_ \texttt{b}_1 \_ \texttt{b}_2 \_ \texttt{b}_3.
\end{equation}
From the names the respective vertices or rather the respective triangle products can be identified.

\subsection{Datasets}
\subsubsection[TOSCA]{TOSCA~\footnote{\url{https://keeper.mpdl.mpg.de/f/cf736ce0e16d4323a13b/?dl=1}}}
A sample generated matching problems (LP-files) on TOSCA~\cite{bronstein2008} shapes. 
The dataset contains $75$ ILP files consisting of problems with $200,000$ binary variables up to $5,000,00$ binary variables.
The dataset contains matching problems between complete shapes and partial shapes and shapes with equal triangulation. Partial shapes are shapes which are missing parts of the original shape (such files are denoted with \texttt{\_partial}). Additionally, there are matching problems which are between non-isometric deformed shapes e.g. a cat matched with a dog (such files are denoted with \texttt{\_noniso}).
\\
The original shapes are not included.
\subsubsection[Deep Learning Predictor Energies]{Deep Learning Predictor Energies~\footnote{\url{https://keeper.mpdl.mpg.de/f/3988e4dd09ce48438a79/?dl=1}}}
A subset of shape matching problems (LP-files) containing in total $167$ ILP files consisting of problems with $100$ up to $900$ triangles in increments of $100$. This translates to $200,000$ up to $14,000,000$ binary variables.
Shape pairs for matching are taken from FAUST\_r~\cite{ren2018continuous,donati2020deep}, SMAL~\cite{zuffi20173d}, SHREC’20~\cite{dyke2020track} and DT4D-H~\cite{magnet2022smooth}. The energies are computed as in~\cite{roetzer2023fast} using the shape descriptors from~\cite{cao2023unsupervised}.

\subsection{Algorithms}
\begin{description}	
	\item[LP + incremental fixation~\cite{windheuser2011geometrically}:]
	LP relaxation of~\eqref{eq:shape-matching-opt} is solved with CPLEX~\cite{cplex} and franctional variables are incrementally fixed towards $\{0,1\}$-values they are close to.
	\item[Eckstein-Bertsekas~\cite{windheuser2011large}:]
	A GPU implementation of the parallelizable primal-dual algorithm proposed by Eckstein and Bertsekas~\cite{eckstein1990alternating}.
\item[SMcomb~\cite{roetzer2022scalable}:] Shape matchings are constructed by iterative fixing + propagation and backtracking algorithm using dual information computed by~\cite{lange2021efficient} as variable guidance.
    \item[FastDOG-LBFGS + Constraint Splitting~\cite{roetzer2023fast}:]
        LP relaxation is solved with an extension of the GPU solver~\cite{abbas2022fastdog} that incorporates LBFGS for faster convergence and splitting long constraints into shorter ones for faster iterations. Rounding is performed with incremental perturbation until the dual produces a consistent primal solution as in~\cite{abbas2022fastdog}.
\end{description}

\clearpage
\section{Discrete Tomography}

The discrete tomography problem is the reconstruction of integral values given by a set of tomographic projections.
It specializes the conventional tomographic reconstruction problem that allows for fractional values.

\begin{definition}[Discrete Tomography]
Given an MRF $(V,E,(\mathcal{L}_v)_{v \in V},(\theta_v)_{v \in V}, (\theta_{uv})_{uv \in E})$ such that the label space is $\mathcal{L}_v = \{0,1,\ldots,K-1\}$ $\forall v \in V$ and a number of projections
$(P_i \subset V, p_i \in \N)_{i=1,\ldots,k}$ the discrete tomography problem is
\begin{equation}
\begin{array}{rl}
\min\limits_{x \in \prod_{v \in V} \mathcal{L}_v} & \sum\limits_{v \in V} \theta_v(x_v) + \sum\limits_{uv \in E} \theta_{uv}(x_u,x_v) \\
\text{s.t.}
& \sum\limits_{v \in P_i} x_v = p_i \quad \forall i \in [k]
\end{array}
\end{equation}
\end{definition}
Typically we have $\theta_v(x_i) = 0$ $\forall v \in V, x_i \in \mathcal{L}_v$, that is the unary potentials do not give any preference.

\begin{figure}[H]
\begin{center}
\includegraphics[width=0.7\columnwidth]{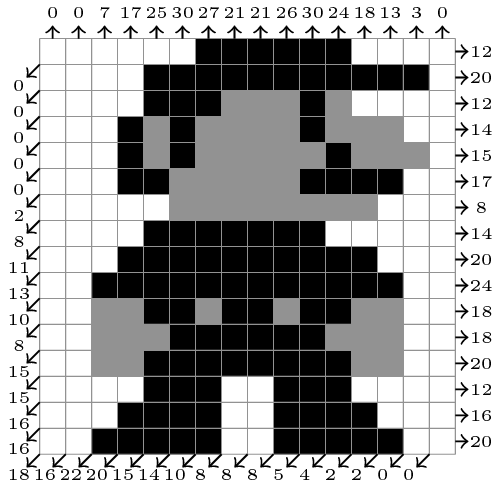}
\end{center}
\caption{
Exemplary discrete tomography problem.
White pixels indicate value $0$, gray ones $1$ and the black ones $2$.
Each tomographic projection is indicated by an arrow and contains those pixels $P_i$ along its ray. The numbers correspond to $p_i$.
}
\label{fig:discrete-tomo}
\end{figure}

An illustration of tomographic projections is given in Figure~\ref{fig:discrete-tomo}.

\subsection{File Format}
We use an extension of the UAI file format used for MRFs in Section~\ref{sec:mrf-file-format}. In addition to specifying the  MRF structure the tomographic projections are appended.

\begin{fileformat}
UAI file format for MRF

PROJECTIONS
.
.
.
(*$i_1$*) + ... (*$i_{\abs{P_i}}$*) =
    (Inf,...,Inf,(*$\underbrace{0}_{p_i\text{-th place}}$*),Inf,...,Inf)
.
.
.
\end{fileformat}
for $P_i = (i_1,\ldots,i_{\abs{P_i}})$ and $i=1,\ldots,K$.

\subsection{Datasets}
\subsubsection[Synthetic Discrete Tomography]{Synthetic Discrete Tomography\footnote{\url{https://keeper.mpdl.mpg.de/f/8827141df8254eefbac4/?dl=1}}}
Around $2700$ discrete tomography instances computed from $30 \times 30$ synthetically generated images with 3 discrete intensity values and varying density of observed objects measured by varying numbers of tomographic projections.
Additionally larger discrete tomography instances of the ``Logan'' image are given.

\subsection{Algorithms}
\begin{description}[style=unboxed]
\item[Subgradient ascent on submodular MRF \& Projection~\cite{kappes2015tomogc}:]
    A binary discrete tomography solver using a Lagrange decomposition into submodular binary MRF solved with graph cuts and tomographic projection constraint solved via linear algebra.
\item[First-order optimization~\cite{zisler2016non}:]
    Combination of total variation regularized reconstruction problem with a non-convex discrete constraint optimized with forward-backward splitting.
\item[Fixed-point iteration~\cite{zisler2016discrete}:]
    Combination of a non-local projection constraint problem with a continuous convex relaxaton of the multilabeling problem solved by a fixed point iteration each of which amounts to solution of a convex auxiliary problem.
\item[Subgradient on decomposition into chain subproblems~\cite{kuske2017novel}:]
    Decompose original problems into chain subproblems that are recursively solved with fast (max,sum)-algorithms.
    The resulting Lagrange decomposition is optimized with a bundle solver.
\item[FW-Bundle Method~\cite{swoboda2019map}:]
    Similar to~\cite{kuske2017novel} but use a Frank-Wolfe based bundle method for optimization of the Lagrange decomposition.
\end{description}

\clearpage
\section{Bottleneck Markov Random Fields}
The bottleneck Markov Random Field problem is an extension of the ordinary Markov Random Field inference problem from Section~\ref{sec:mrf}.
The ordinary MRF problem is an optimization w.r.t.\ the $(\min,+)$-semiring (i.e.\ we minimize over a sum of potentials), while the inference over bottleneck MRFs is additional an optimization w.r.t.\ the $(\min,\max)$-semiring (i.e.\ we also optimize over the maximum assignment of all potentials).

\begin{definition}[Bottleneck MRF]
\label{def:bottleneck-mrf}
Given an MRF $(V,E,\mathcal{L},\theta)$ (see Defintion~\ref{def:mrf}) and additional bottleneck potentials $\psi_v : \mathcal{L}_v \rightarrow \R$ for $v \in V$ and $\psi_{uv} : \mathcal{L}_u \times \mathcal{L}_v \rightarrow \R$ for $uv \in E$, the bottleneck MRF inference problem is
\begin{multline}
\min_{x \in \mathcal{L}}  \sum_{v \in V} \theta_v(x_v) + \sum_{uv \in E} \theta_{uv}(x_u,x_v) + \\ \min\{\max_{v \in V} \psi(x_v), \max_{uv \in E} \psi_{uv}(x_u,x_v) \}
\end{multline}
\end{definition}

\subsection{File Format}
The file format is an extension of the uai file format proposed for MRFs in Section~\ref{sec:mrf-file-format}.

\begin{fileformat}
MARKOV
(*$\theta \text{ potentials in uai format as in Section~\ref{sec:mrf-file-format}}$*)

MAX-POTENTIALS
(*$\psi \text{ potentials in uai format as in Section~\ref{sec:mrf-file-format}}$*)
\end{fileformat}
What follows after MARKOV is the ordinary MRF part describing $\theta$ potentials in Definition~\ref{def:bottleneck-mrf} and what follows after MAX-POTENTIALS are the $\psi$ potentials for the bottleneck part in Definition~\ref{def:bottleneck-mrf}.

\subsection{Datasets}
\subsubsection[Horizon Tracking]{Horizon Tracking\footnote{https://keeper.mpdl.mpg.de/f/4b0af028e879482b8037/?dl=1}}
12 horizon tracking instances and ground truth data for tracking subsurface rock layers generated from subsurface volumes F3 Netherlands, Opunake-3D, Waka-3D from The Society of Exploration Geophysicists.

\subsection{Algorithms}
\begin{description}[style=unboxed]
\item[Lagrange Decomposition \& Subgradient Ascent~\cite{abbas2019bottleneck}:] Combinatorial subproblems for ordinary tree MRFs and bottleneck chain MRFs joined via Lagrange multipliers and optimized via subgradient ascent.
\end{description}

\clearpage

{\small
\bibliographystyle{alpha}
\bibliography{literature}
} 


\end{document}